\documentclass{article}

\PassOptionsToPackage{numbers, compress}{natbib}
\usepackage[preprint]{tencent_hunyuan}

\usepackage[utf8]{inputenc} 
\usepackage[T1]{fontenc}    
\usepackage{hyperref}       
\usepackage{url}            
\usepackage{booktabs}       
\usepackage{amsfonts}       
\usepackage{nicefrac}       
\usepackage{microtype}      
\usepackage{xcolor}         
\usepackage{amsmath}
\usepackage{graphicx} 
\usepackage{multirow}
\usepackage{wrapfig}

\usepackage{subcaption} 

\newcommand{\name}{Hunyuan-DiT~}
\newcommand{\namens}{Hunyuan-DiT}

\title{\name: A Powerful Multi-Resolution Diffusion Transformer with Fine-Grained Chinese Understanding}

\author{
Zhimin Li$^*$,  
Jianwei Zhang\thanks{ equal contribution},~~
Qin Lin,  
Jiangfeng Xiong,  
Yanxin Long,  
Xinchi Deng,  
Yingfang Zhang, \AND  
Xingchao Liu,  
Minbin Huang,  
Zedong Xiao,  
Dayou Chen,  
Jiajun He,  
Jiahao Li,  
Wenyue Li,  
Chen Zhang,  \AND
Rongwei Quan,  
Jianxiang Lu,  
Jiabin Huang,  
Xiaoyan Yuan,  
Xiaoxiao Zheng,  
Yixuan Li,  
Jihong Zhang,  \AND
Chao Zhang,  
Meng Chen,  
Jie Liu,  
Zheng Fang,  
Weiyan Wang,  
Jinbao Xue,  
Yangyu Tao,  
Jianchen Zhu,  \AND
Kai Liu,  
Sihuan Lin,  
Yifu Sun,  
Yun Li,  
Dongdong Wang,  
Mingtao Chen,
Zhichao Hu,  
Xiao Xiao,  
Yan Chen,  \AND 
Yuhong Liu, 
Wei Liu,  
Di Wang,  
Yong Yang,  
Jie Jiang,  
Qinglin Lu\thanks{ corresponding author (Email: \url{qinglinlu@tencent.com})} \\
\\
\text{Tencent Hunyuan}
}

\begin{document}

\maketitle

\begin{abstract}
We present \namens, a text-to-image diffusion transformer with fine-grained understanding of both English and Chinese. To construct \namens, we carefully design the transformer structure, text encoder, and positional encoding. We also build from scratch a whole data pipeline to update and evaluate data for iterative model optimization. For fine-grained language understanding, we train a Multimodal Large Language Model to refine the captions of the images. Finally, \name can perform multi-turn multimodal dialogue with users, generating and refining images according to the context.
Through our holistic human evaluation protocol with more than 50 professional human evaluators, \name sets a new state-of-the-art in Chinese-to-image generation compared with other open-source models.
Code and pretrained models are publicly available at~\url{github.com/Tencent/HunyuanDiT}
\end{abstract}

\section{Introduction}

\begin{figure}
    \centering
    \vspace{-15pt}
    \includegraphics[width=0.8\textwidth]{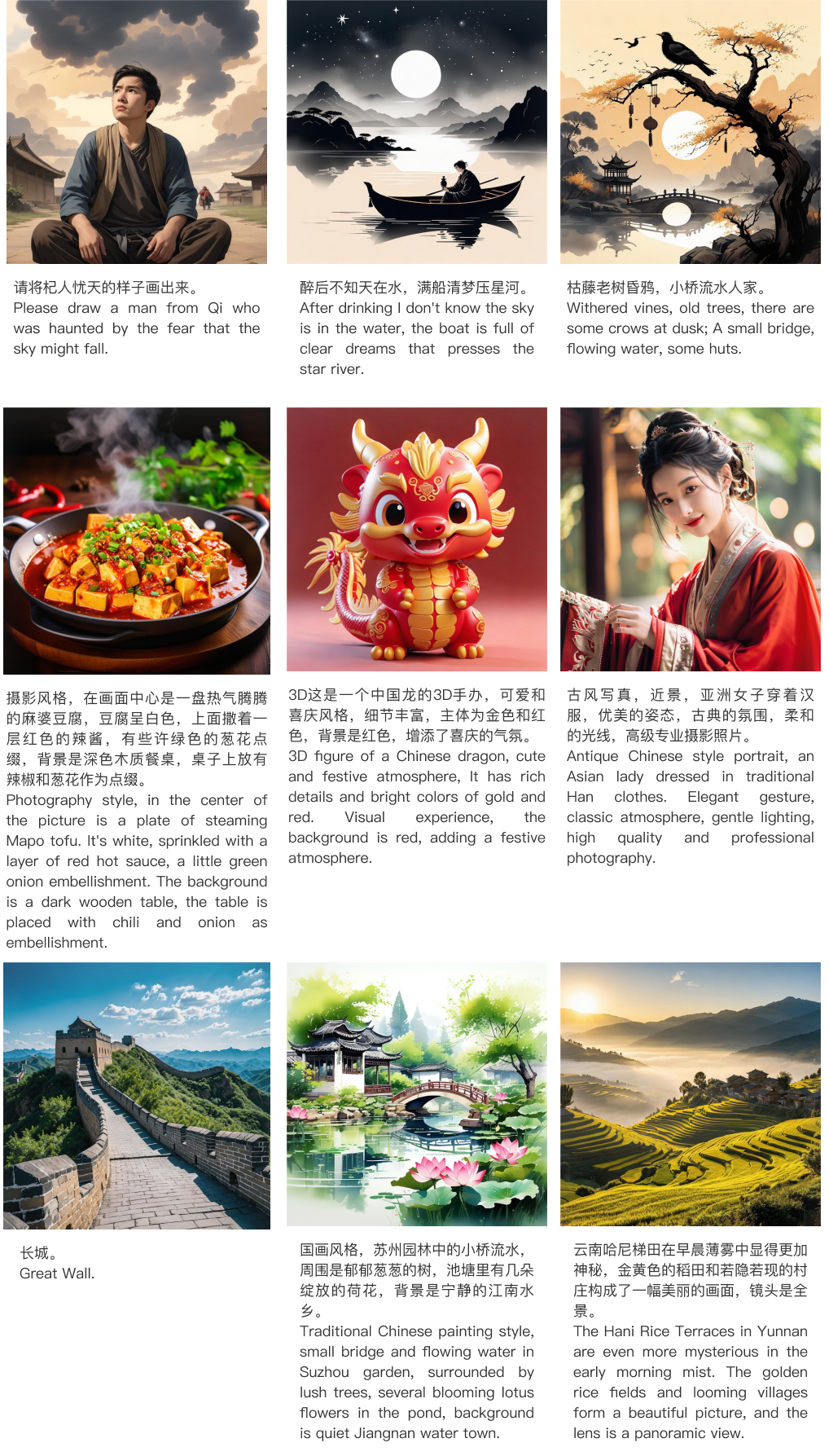}
    \vspace{-10pt}
    \caption{\name can generate images containing Chinese elements. In this report, without further notice, all the images are directly generated using Chinese prompts.}
    \label{fig:chinese}
\end{figure}

\begin{figure}
    \centering
    \includegraphics[width=0.8\textwidth]{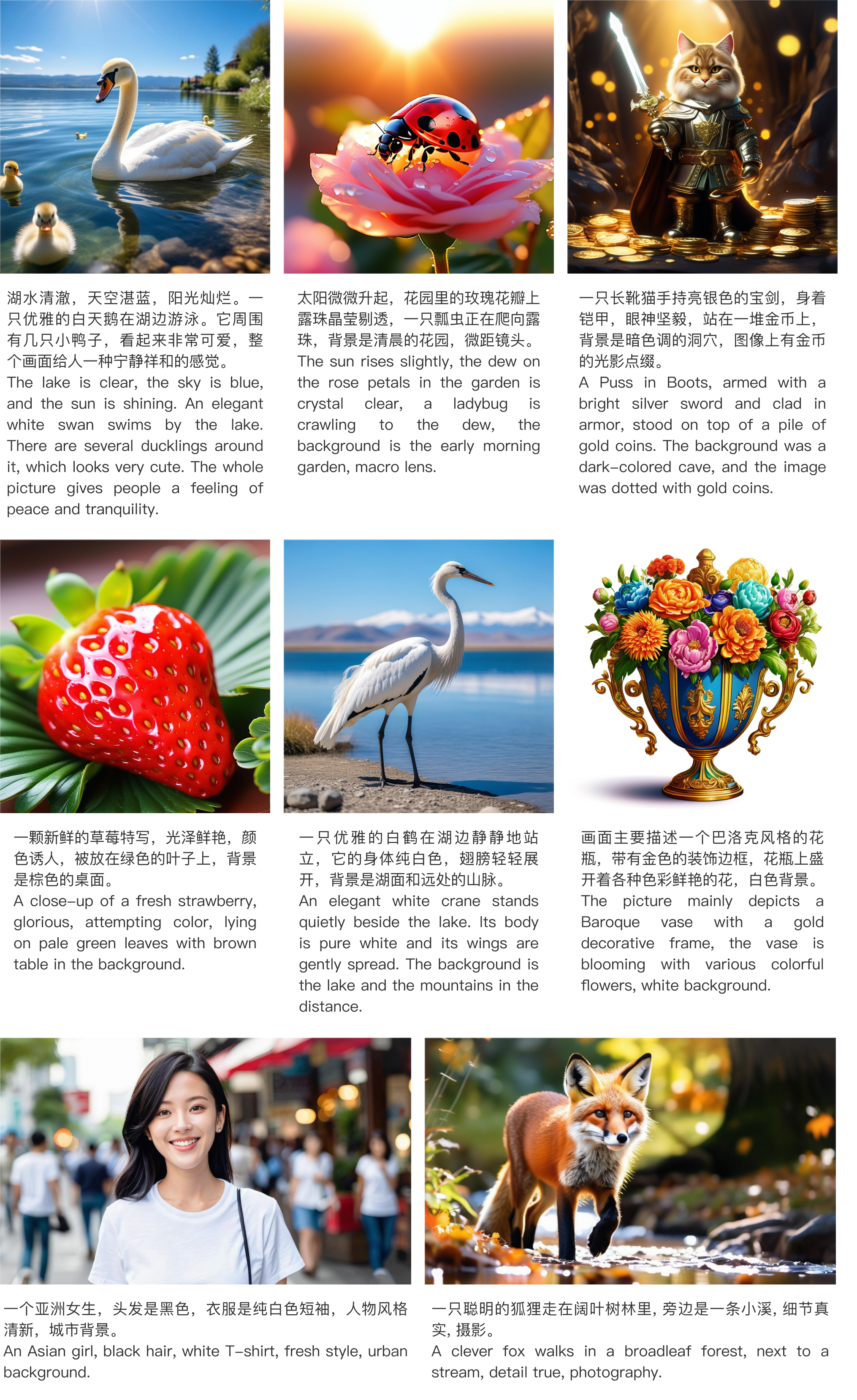}
    \caption{\name can generate images according to fine-grained text prompts.}
    \label{fig:finegrain}
\end{figure}

\begin{figure}
    \centering
    \includegraphics[width=0.8\textwidth]{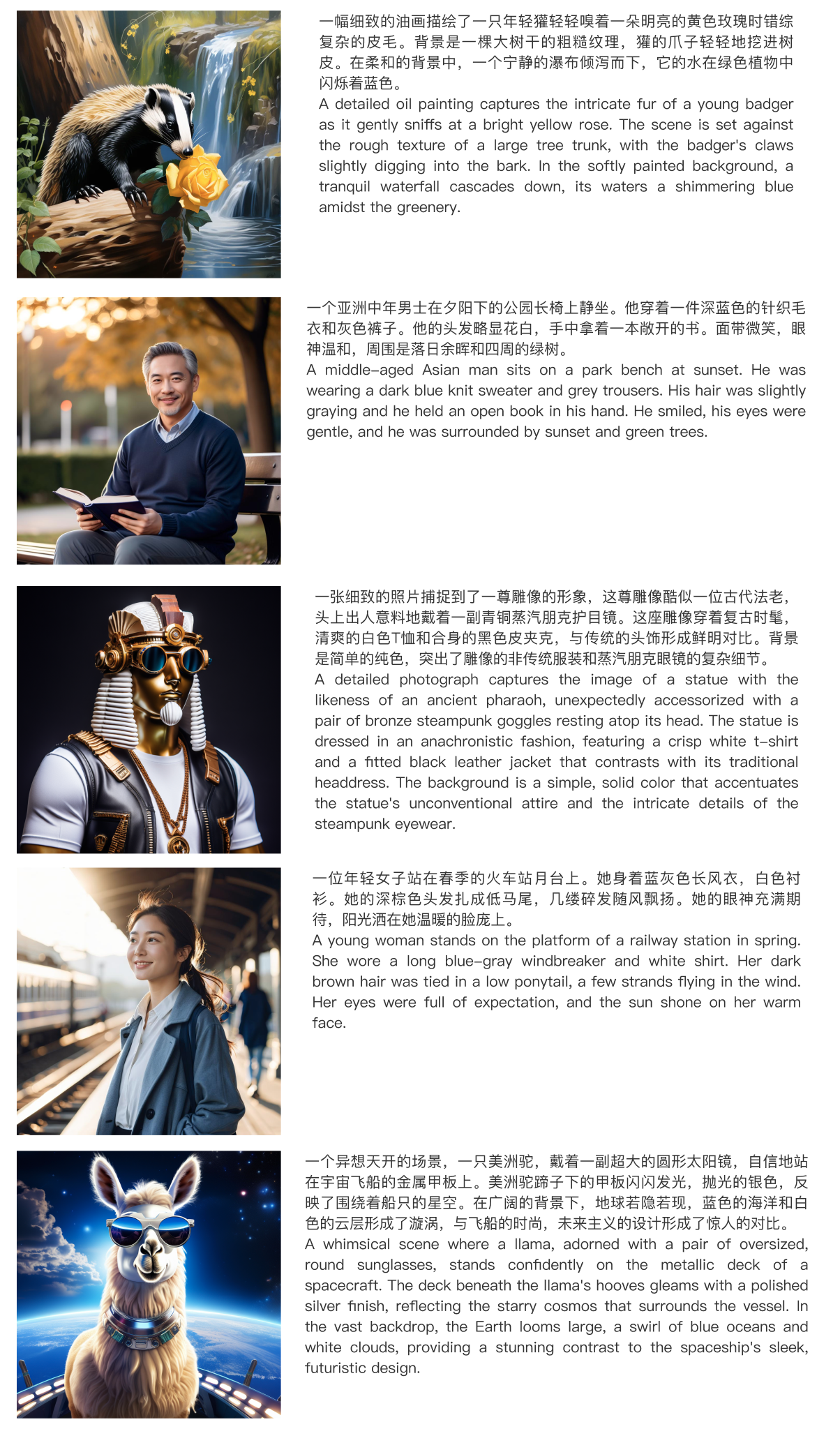}
    \caption{\name can generate images following long text prompts.}
    \label{fig:longtext}
\end{figure}

\begin{figure}
    \centering
    \includegraphics[width=0.8\textwidth]{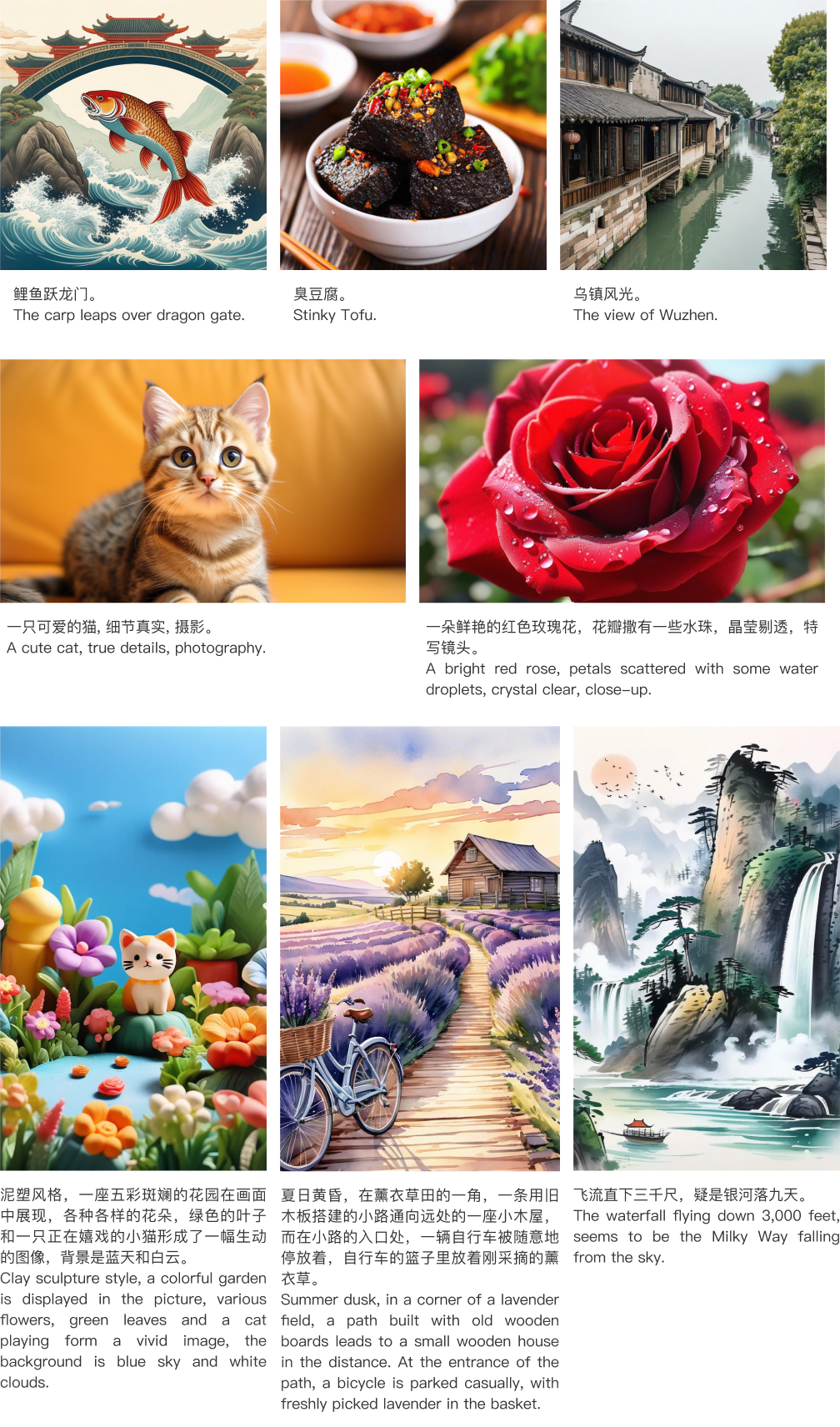}
    \caption{\name can generate images in various resolutions.}
    \label{fig:multires}
\end{figure}

\begin{figure}
    \centering
    \includegraphics[width=0.8\textwidth]{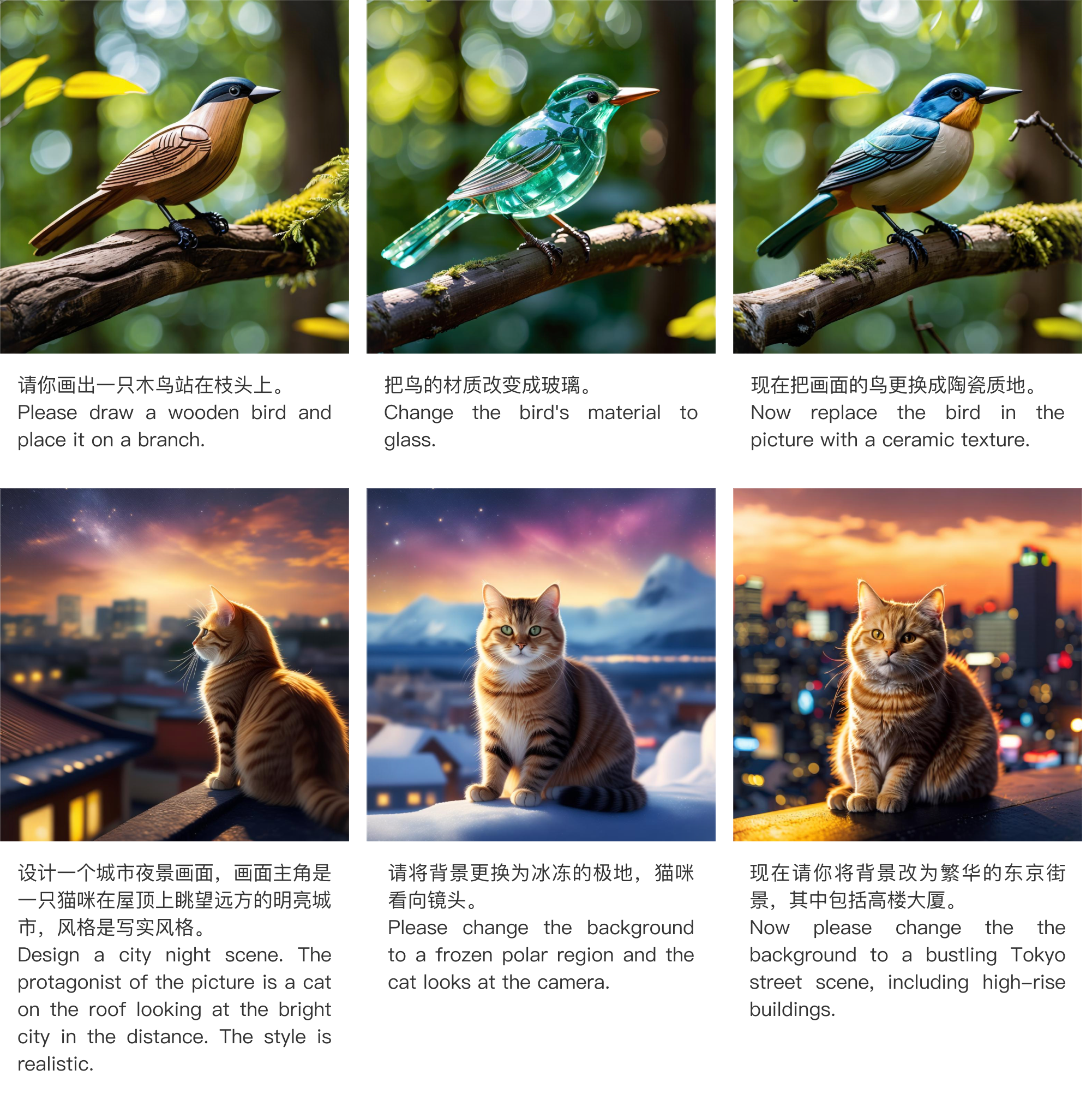}
    \caption{\name can generate images in multi-turn dialogue.}
    \label{fig:multiturn}
\end{figure}

Diffusion-based text-to-image generative models, such as DALL-E~\citep{betker2023improving}, Stable Diffusion~\citep{podell2023sdxl, esser2024scaling} and Pixart~\citep{chen2023pixart}, have shown the ability to generate images with unprecedented quality.
However, they lack the ability to directly understand Chinese prompts, limiting their potential in image generation with Chinese text prompts.
To improve Chinese understanding, AltDiffusion~\citep{ye2024altdiffusion}, PAI-Diffusion~\citep{wang2023pai} and Taiyi~\citep{wu2024taiyi} were proposed but their generation quality still needs improvement.

In this report, we introduce our entire pipeline for constructing \name, which can generate detailed high-quality images in multiple different resolutions according to both English and Chinese prompts. \name is made possible by our following efforts: 
(1) we design a new network architecture based on diffusion transformer~\citep{peebles2023scalable}. It combines two text encoders, a bilingual CLIP~\citep{radford2021learning} and a multilingual T5 encoder~\citep{raffel2020exploring} to improve language understanding and increase the context length.
(2) we build from scratch a data processing pipeline to add data, filter data, maintain data, update data and apply data to optimize our text-to-image model. Specifically, an iterative procedure called `data convoy' is designed to examine the effectiveness of new data.
(3) we refine the raw captions in the image-text data pairs with Multimodal Large Language Model (MLLM). Our MLLM is fine-tuned to generate structural captions with world knowledge.
(4) we enable \name to interactively modify its generation by having multi-turn dialogues with the user.
(5) we perform post-training optimization in the inference stage to lower the deployment cost of \name. 

To thoroughly evaluate the performance of \name, we also created an evaluation protocol with $\geq 50$ professional evaluators. The protocol carefully takes into account the different dimensions of a text-to-image model, including text-image consistency, AI artifacts, subject clarity, aesthetics, etc. Our evaluation protocol is incorporated into the data convoy to update the generative model.

Our model, \name, achieves state-of-the-art performance among open-source models. In Chinese-to-image generation, \name is the best in text-image consistency, excluding AI artifacts, subject clarity, and aesthetics compared with existing open-source models, including Stable Diffusion 3. It performs similarly as top closed-source models, such as DALL-E 3 and MidJourney v6, in subject clarity and aesthetics. Qualitatively, for Chinese elements understanding, including categories such as ancient Chinese poetry and Chinese cuisine, \name can generate results with higher image quality and semantic accuracy compared to other comparison algorithms. \name supports long text understanding up to 256 tokens. 
\name can generate images using both Chinese and English text prompts. In this report, without further notice, all the images are generated using Chinese prompts. 
\section{Methods}

\subsection{Improved Generation with Diffusion Transformers}
\begin{figure}[t]
    \centering
    \includegraphics[width=0.8\textwidth]{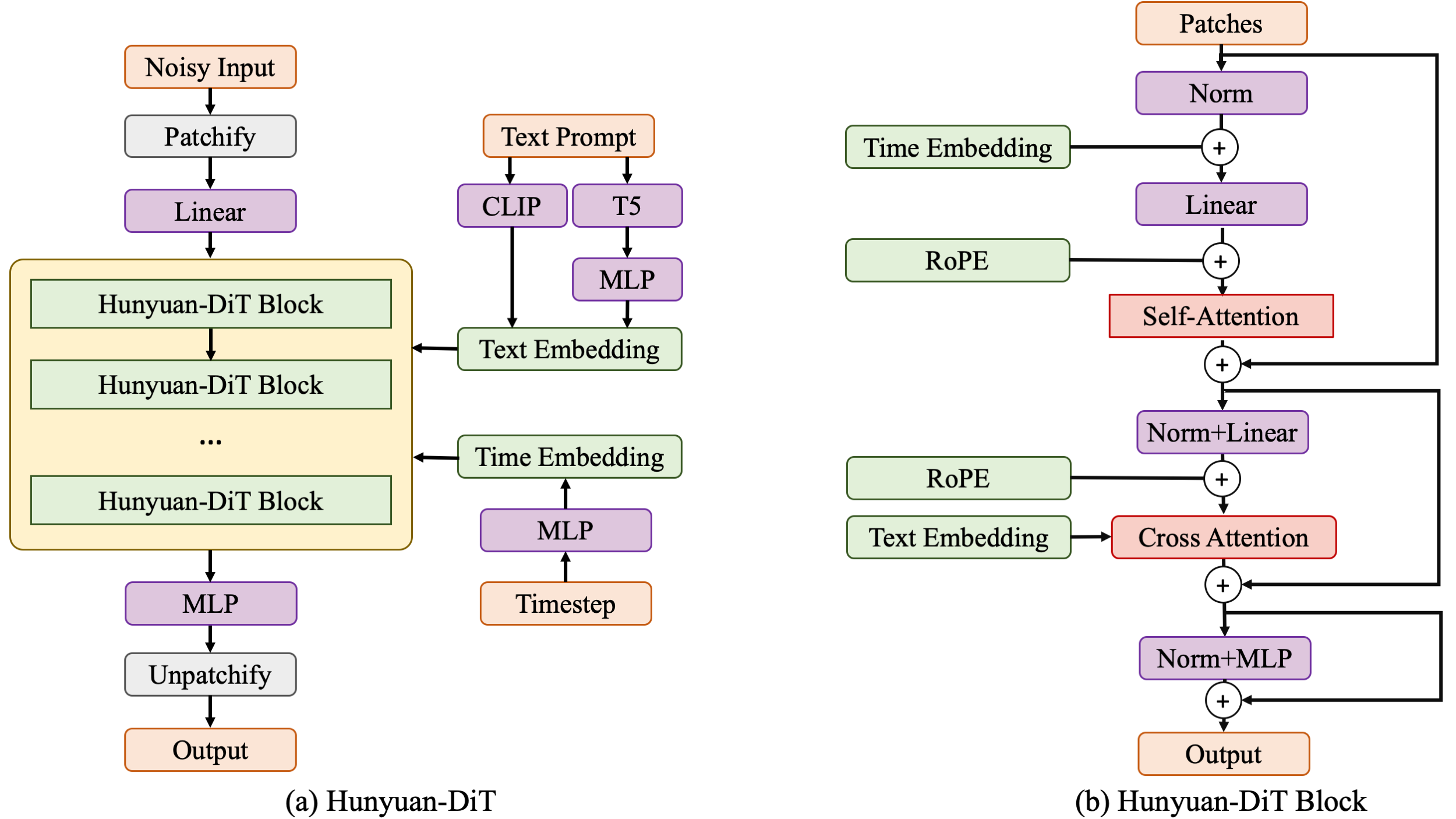}
    \caption{The model structure of \name. }
    \label{fig:model_strucutre}
\end{figure}

\name is a diffusion model in the latent space, as depicted in Figure~\ref{fig:model_strucutre}. Following the Latent Diffusion Model~\citep{rombach2022high}, we use a pre-trained Variational Autoencoder (VAE) to compress the images into low-dimensional latent spaces and train a diffusion model to learn the data distribution with diffusion models. Our diffusion model is parameterized with a transformer~\citep{vaswani2017attention, peebles2023scalable, bao2023all}. To encode the text prompts, we leverage a combination of pre-trained bilingual (English and Chinese) CLIP~\citep{radford2021learning} and multilingual T5 encoder~\citep{raffel2020exploring}. We will introduce the details of each module in sequel.

\paragraph{VAE} We use the VAE in SDXL~\citep{podell2023sdxl}, which is fine-tuned on 512 $\times$ 512 images from the VAE in SD 1.5~\citep{rombach2022high}. 
Experimental findings show that the text-to-image models trained on the high-resolution SDXL VAE improved clarity, alleviated over-saturation, and reduced distortions over SD 1.5 VAE.
As the VAE latent space greatly influences generation quality, we will explore a better training paradigm for the VAE in the future.

\paragraph{The Diffusion Transformer in \name}
Our diffusion transformer has several improvements compared to the baseline DiT~\citep{peebles2023scalable}. We found the Adaptive Layer Norm used in class-conditional DiT performs unsatisfactorily to enforce fine-grained text conditions. Therefore, we modify the model structure to combine the text condition with the diffusion model using cross-attention as Stable Diffusion~\citep{rombach2022high}. 
\name takes a vector $x \in \mathbb{R}^{c \times h \times w}$ in the latent space of the VAE as input, and then patchifies $x$ into $\frac{h}{p} \times \frac{w}{p}$ patches, where $p$ is set to 2. After a linear projection layer, we get $hw / 4$ tokens for the subsequent transformer blocks. 
\name has two types of transformer blocks, the encoder block and the decoder block. Both of them contain three modules - self-attention, cross-attention, and feed-forward network (FFN). The text information is fused in the cross-attention module. The decoder block additionally contains a skip module, which adds the information from the encoder block in the decoding stage. The skip module is similar to the long skip-connection in U-Nets, but there are no upsampling or downsampling modules in \name due to our transformer structure. 
Finally, the tokens are reorganized to recover the two-dimensional spatial structure.
For training, we find using v-prediction~\citep{salimans2021progressive} gives better empirical performance.

\paragraph{Text Encoder} 
\begin{wrapfigure}{r}{0.4\textwidth}
\vspace{-15pt}
  \begin{center}
    \includegraphics[width=0.38\textwidth]{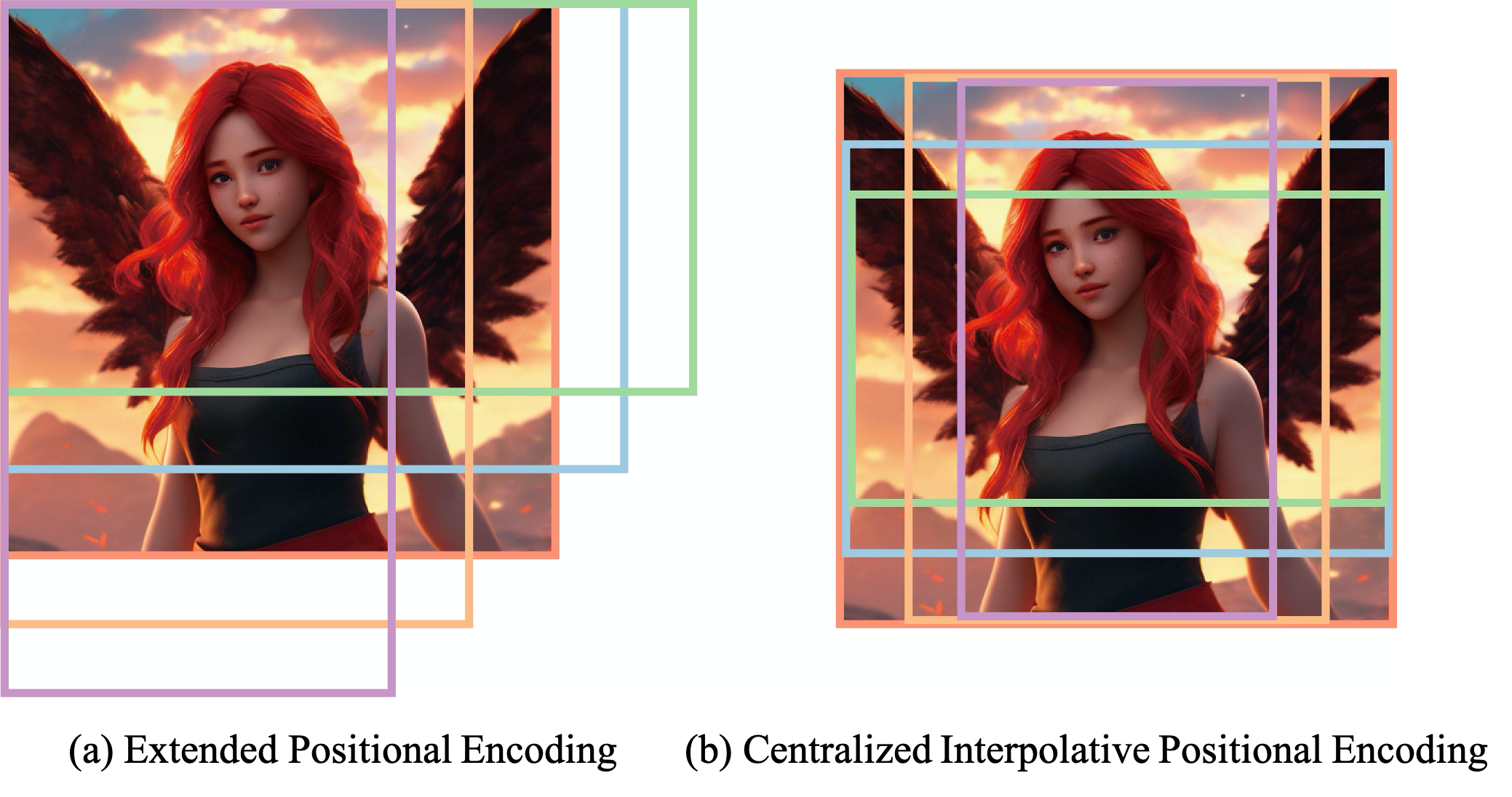}
  \end{center}
  \vspace{-10pt}
  \caption{Illustration of Extended Positional Encoding and Centralized Interpolative Positional Encoding}
  \vspace{-25pt}
\end{wrapfigure}
An efficient text encoder is crucial in text-to-image generation, as they need to accurately understand and encode the input text prompts to generate corresponding images. CLIP~\citep{radford2021learning} and T5~\citep{raffel2020exploring} have become the mainstream choices for these encoders. Matryoshka diffusion models~\citep{gu2023matryoshka}, Imagen~\citep{saharia2022photorealistic}, MUSE~\citep{chang2023muse}, and Pixart-$\alpha$~\citep{chen2023pixart} use solely T5 to enhance their understanding of the input text prompts. 
In contrast, eDiff-I~\citep{balaji2022ediff} and Swinv2-Imagen~\citep{li2023swinv2} fuse the two encoders, CLIP and T5, to further improve their text understanding capabilities.
\name chooses to combine T5 and CLIP in text encoding to leverage the advantages of both models, thereby enhancing the accuracy and diversity of the text-to-image generation process.

\paragraph{Positional Encoding and Multi-Resolution Generation}
A common practice in visual transformers~\citep{peebles2023scalable, dosovitskiy2020image} is to apply sinusoidal positional encoding that encodes the absolute position of a token. 
In \name, we employ the Rotary Positional Embedding (RoPE)~\citep{su2024roformer} to simultaneously encode the absolute position and relative position dependency.
We use two-dimensional RoPE which extends RoPE to the image domain. 

\name supports multi-resolution training and inference, which requires us to assign appropriate positional encodings for different resolutions. For $x \in \mathbb{R}^{c \times h \times w}$,  we tried two types of positional encoding for multi-resolution generation:
\begin{enumerate}
    \item \textbf{Extended Positional Encoding:} Extended Positional Encoding gives the positional encoding of $x$ in a naive way, which is,
    \begin{equation}
        \text{PE}(x_{i,j}) = \left(f(i), f(j)\right),~~~i \in \{1,\cdots, h\}, j \in \{1,\cdots,w\},
    \end{equation}
    where $f$ is the positional encoding function for each coordinate $i$ and $j$. $\text{PE}(x)$ is the obtained 2D positional encoding for the position $(i, j)$. Note that when the data $x$ has different resolutions, their $h$ and $w$ exhibit huge differences and the positional encoding varies significantly.
    \item \textbf{Centralized Interpolative Positional Encoding:} We use Centralized Interpolative Positional Encoding to align the positional encoding for $x$ with different $h$ and $w$. Assuming $h \geq w$, Centralized Interpolative Positional Encoding computes the positional encoding as, 
    \begin{equation}
        \text{PE}(x_{i,j}) = \left(f \left ( \frac{S}{2} +  \frac{S}{h}  \left ( i - \frac{h}{2} \right) \right),  f \left ( \frac{S}{2} +  \frac{S}{h}  \left ( j - \frac{w}{2} \right) \right)  \right), 
    \end{equation}
    where $i \in \{1,\cdots, h\}, j \in \{1,\cdots,w\}$ and $S$ is a pre-defined boundary of the positional encoding. This strategy ensures images with various resolutions to have the same range $[0, S]$ when computing positional encoding, therefore improving the efficiency of learning.
\end{enumerate}

Although Extended Positional Encoding is easier to implement, we observe that it is a suboptimal choice for multi-resolution training. It could not align images with different resolutions or cover the rare cases where both $h$ and $w$ are large. On the contrary, Centralized Interpolative Positional Encoding allows images with different resolutions to share similar positional encoding spaces. With Centralized Interpolative Positional Encoding, the model converges faster and generalizes to new resolutions.

\paragraph{Improving Training Stability}
To stabilize training, we present three techniques:
\begin{enumerate}
    \item We add layer normalization in all the attention modules before computing Q, K, and V. This technique is called QK-Norm, which is proposed in~\citep{henry2020query}. We found it effective for training \name as well.
    \item We add layer normalization after the skip module in the decoder blocks to avoid loss explosion during training.
    \item We found certain operations, e.g., layer normalization, tend to overflow with FP16. We specifically switch them to FP32 to avoid numerical errors.
\end{enumerate}

\subsection{Data Pipeline}

\paragraph{Data Processing} The pipeline to preparing our training data is composed of four parts, which is illustrated in Fig.\ref{fig:data_pipeline}

\begin{enumerate}
    \item \textbf{Data Acquisition:} The primary channels for data acquisition are currently external purchasing, open data downloading, and authorized partner data.
    \item \textbf{Data Interpretation:} After obtaining the raw data, we tag the data to identify the strengths and weaknesses of the data. Currently, over ten tagging capabilities are supported, including image clarity, aesthetics, indecency, violence, sexual content, presence of watermarks, image classification, and image description.
    \item \textbf{Data Layering}: Data layering is constructed for large quantities of images to serve the different stages of model training. For example, billions of image-text pairs are used as copper-tier data to train our foundational CLIP model~\citep{radford2021learning}. Then, a relatively high-quality image set is screened from this large library as silver-tier data to train the generative model to improve the model's quality and understanding capabilities. Lastly, through machine screening and manual annotation, the highest quality data is selected as gold-tier data for refining and optimizing the generative model.
    \item \textbf{Data Application}: The hierarchical data are applied to several areas. Specialized data is filtered out for specialty optimizations, e.g, person or style specializations. Newly processed data is continually added to the iterative optimization of the foundation generative model. The data is also frequently inspected to maintain the quality of the ongoing data processing.
\end{enumerate}

\paragraph{Data Category System}
We found the coverage of the data categories in the training data crucial for training accurate text-to-image models. Here we discuss two fundamental categories:
\begin{enumerate}
    \item \textbf{Subject:} The generation of the subject is the foundational ability of the text-to-image model. Our training data covers a vast majority of categories, including human, landscape, plants, animals, goods, transportation, games, and more, with over ten thousand sub-categories.
    \item \textbf{Style:} The diversity of the style is critical to the user's preference and stickiness. Currently, we have covered over a hundred styles, including anime, 3D, painting, realistic, and traditional styles.
\end{enumerate}

\paragraph{Data Evaluation}
To evaluate the impact of introducing specialized data or newly processed data on the generative model, we design a `data convoy' mechanism, depicted in Fig.\ref{fig:data_convoy}, which is composed of:
\begin{enumerate}
    \item 
    We categorize the training data according to the data category system, containing subject, style, scene, composition, etc. Then we adjust the distribution between different categories to meet the model's demand and fine-tune the model with the category-balanced dataset.
    \item
    We perform category-level comparisons between the fine-tuned model and the original model to evaluate the advantages and drawbacks of the data, relying on which we set the directions to update our data.
\end{enumerate}

Successfully running the mechanism requires a complete evaluation protocol on the text-to-image model. Our model evaluation protocol is composed of two parts: 
\begin{enumerate}
    \item \textbf{Evaluation Set Construction:} We construct the initial evaluation set by combining bad cases and business needs based on our data categories. Through human annotation of the reasonableness, logic, and comprehensiveness of the test cases, the usability of the evaluation set is assured.
    \item \textbf{Evaluation in Data Convoy:} In every data convoy, we randomly select a subset of test cases from the evaluation set to form a holistic evaluation subset including subjects, styles, scenes, compositions. We compute an overall score of all the evaluated dimensions to assist the iteration of data.
\end{enumerate}
We will elaborate our evaluation protocol in Section~\ref{sec:eval}.

\subsection{Caption Refinement for Fine-Grained Chinese Understanding}
The image-text pairs obtained from crawling the Internet are usually low-quality pairs, and improving the corresponding captions for the images is important for training text-to-image models~\citep{chen2023pixart, betker2023improving}. \name adopts a well-trained multimodal large language model (MLLM) to re-caption the raw image-text pairs to enhance the data quality. We adopt structural captions to comprehensively describe the images. Furthermore, we also use raw captions and expert models that include world knowledge to enable the generation of special concepts in the re-captioning.

\paragraph{Re-captioning with Structural Captions} 

Existing MLLMs, e.g., BLIP-2~\citep{li2023blip} and Qwen-VL~\citep{bai2023qwen} tend to generate over-simplified captions that resemble MS-COCO captions~\citep{lin2014microsoft} or highly redundant captions that are not related to the images. To train an MLLM that is suitable to improve raw image-text pairs, we construct a large-scale dataset for structural captions and fine-tune the MLLM.

We use an AI-assisted pipeline for dataset construction. Human labeling for image captioning is difficult, and the labeling quality can hardly be standardized. Therefore, we use a three-stage pipeline to boost labeling efficiency with AI assistance. In Stage 1, we ensemble the captions from multiple basic image captioning models with human labeling to get an initial dataset. In Stage 2, we train the MLLM with the initial dataset, and then use the trained model to generate new captions for the images. As its re-captioning accuracy is enhanced, the efficiency of human labeling is improved by around 4 times. 

Our model structure is similar to LLAVA-1.6~\citep{liu2023improved}. It is composed of a ViT for vision, a decoder-only LLM for language, and an Adapter for bridging vision and text. The training objective is the classification loss as other auto-regressive models.

\paragraph{Re-captioning with Information Injection}
In human labeling of structural captions, world knowledge is always missing because it is impossible for human to recognize all the special concepts in the images.

\begin{figure}[t]
    \centering
    \includegraphics[width=0.9\textwidth]{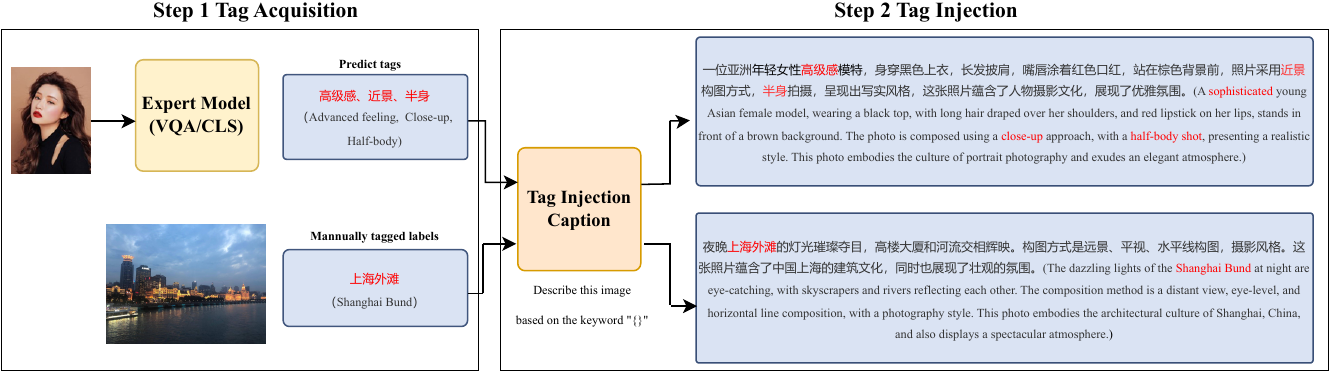}
    \caption{
    Re-captioning with tag injection based on manual labeling and expert models.
    }
    \label{fig:tag_injection_caption}
\end{figure}

We leverage two methods to inject world knowledge to the captions:
\begin{enumerate}
    \item \textbf{Re-captioning with Tag Injection:} To simplify the labeling process, we can label tags of images and use MLLMs to generate tag-injected captions from the labeled tags. Besides labeling with human experts, we can use expert models to get the tags, including but not limited to general object detectors, landmark classification models, and action recognition models. The additional information from tags can significantly add to the world knowledge in the generated captions. To this end, we design an MLLM that takes images and tags as input and outputs more comprehensive captions containing the information from the tags. We found this MLLM can be trained with very sparse human-labeled data.
    \item \textbf{Re-captioning with Raw Captions:} Capsfusion~\citep{yu2023capsfusion} proposed to fuse raw captions with generated descriptive captions using ChatGPT. However, raw captions are usually noisy and LLM alone cannot correct the wrong information in the raw captions. To alleviate this, we construct an MLLM that generates captions from both images and raw captions, which can correct the mistakes by taking image information into account.
\end{enumerate}

\subsection{{Prompt Enhancement with Multi-Turn Dialogue}}

Understanding natural language instructions and performing multi-turn interactions with users are important for a text-to-image system. It can help build a dynamic and iterative creation process that brings the user's idea into reality step by step. In this section, we will detail how we empower \name with the ability to perform multi-turn conversations and image generation.
Various works have made efforts to equip text-to-image models with the multi-turn ability using MLLMs, such as Next-GPT~\citep{wu2023next}, SEED-LLaMA~\citep{ge2023making}, RPG~\citep{yang2024mastering}, and DALLE-3~\citep{betker2023improving}. 
These models either use the MLLM to generate text prompts or the text embeddings for the text-to-image model. We choose the first choice as generating text prompts is more flexible. We train MLLM to understand the multi-turn user dialogue and output the new text prompt for image generation.

\begin{figure}[t]
    \centering
    \includegraphics[width=0.9\textwidth]{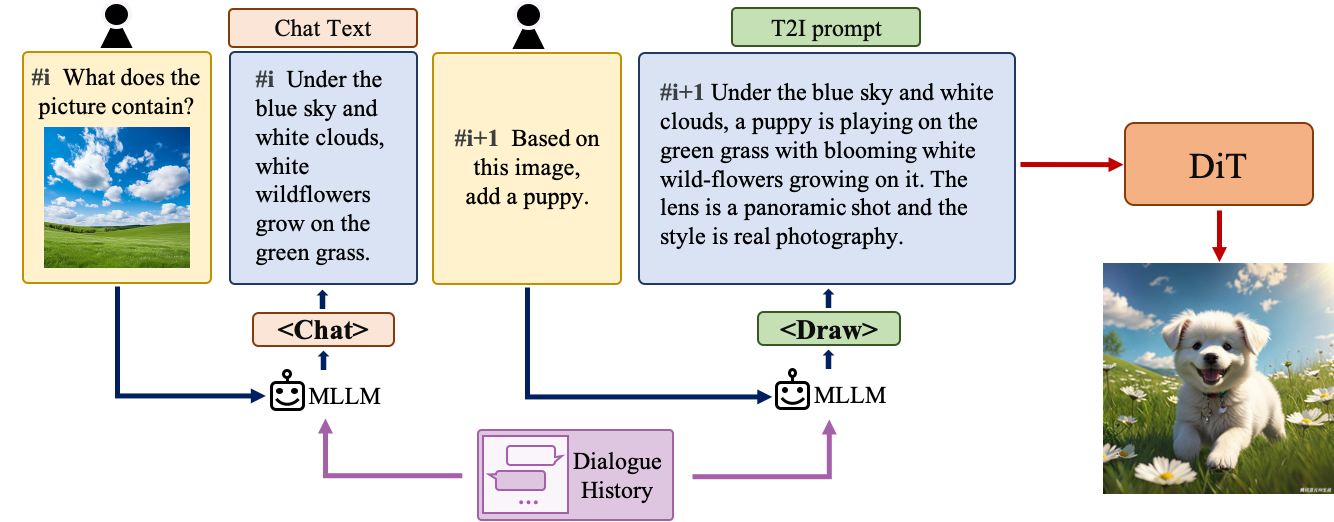}
    \caption{Our pipeline of text-to-image generation with multi-turn dialogue.}
    \label{fig:promptEnhance}
\end{figure}

\paragraph{Text Prompt Enhancement}
Natural language instructions given by the user have a huge difference with the refined captions on which the text-to-image generative model is trained.
Consequently, we need a model to transform these instructions into detailed semantically coherent text prompts for successful high-quality image generation.
To train this model, we use the in-context learning ability of GPT-4. We collect a small set of manually annotated \texttt{(instruction, text prompt)} pairs as in-context learning examples, then we query GPT-4 to generate more data pairs. These pairs construct a single-turn instruction-to-prompt dataset, referred to as $D_p$.

\paragraph{Multimodal Multi-Turn Dialogue} Normal MLLMs only support text output. To align with our goal to build a multi-turn text-to-image generation system, we add a special token \texttt{<draw>} to indicate that a text prompt should be sent to \name in the current turn of conversation.
If the model successfully predicts the \texttt{<draw>} token, it will generate a detailed prompt for \name.
To train the MLLM, we design a dataset of three-turn multimodal conversations. To ensure broad coverage of conversational scenarios, we explore different combinations of input and output types based on four primary categories, i.e., text $\rightarrow$ text, text $\rightarrow$ image, text+image $\rightarrow$ text, text+image $\rightarrow$ image.
By selecting a type in each turn of conversation, we pre-define a set of three-turn dialogue compositions.
For each composition, we then employ GPT-4 to generate the `dialogue prompts', which are used to define the behavior of the AI agent before the dialogue, leading to unique conversational flows. We traverse 13 topics and 7 image editing methods to yield $\sim$15,000 samples after querying GPT-4 with various `dialogue prompts'.
In the `dialogue prompts', we also add the samples in $D_p$ to avoid the distribution shift of the generated text prompts.
We denote this dataset of three-turn text-to-image conversations as $D_{tt}$. 

\begin{figure}[t]
    \centering
    \includegraphics[width=0.9\textwidth]{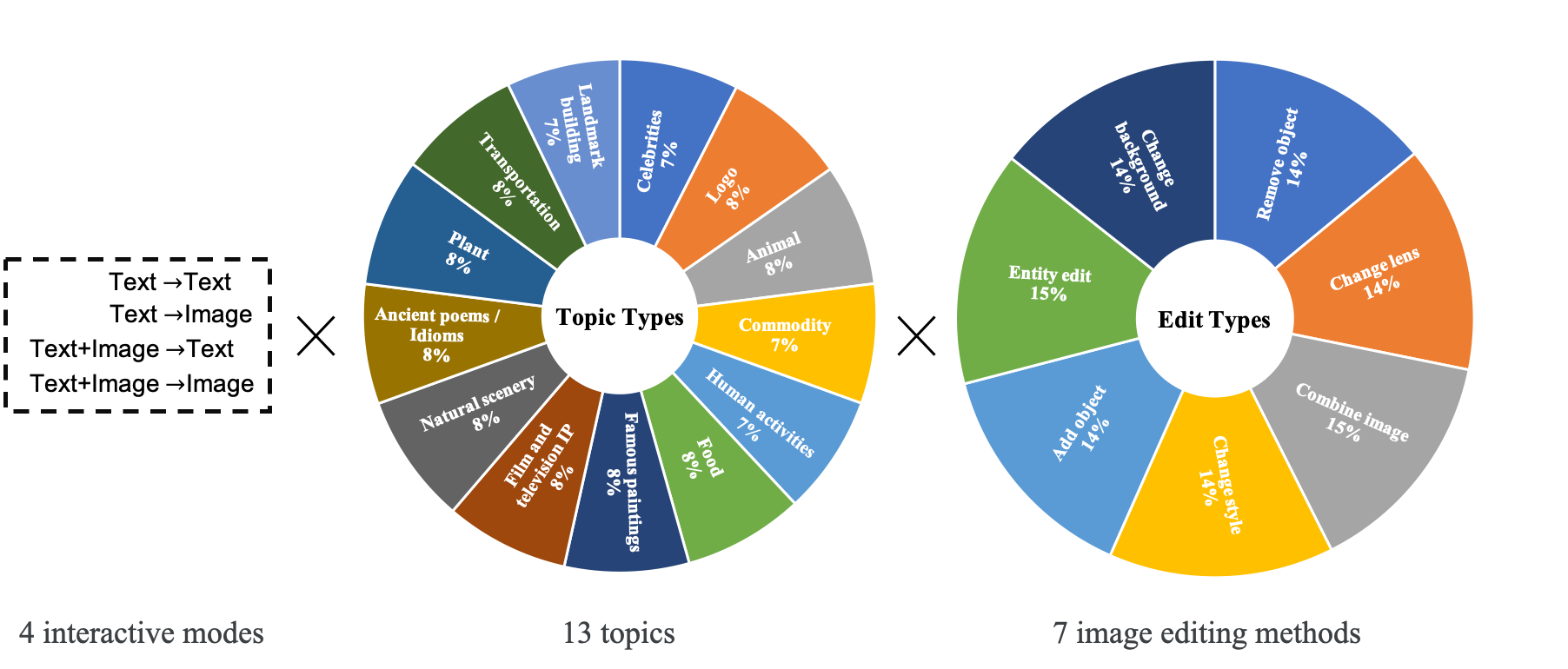}
    \caption{Data construction for multi-turn dialogue.}
    \label{fig:multi_turn_dialogue_data}
\end{figure}

\paragraph{Instruction Tuning Data Mixing}
To maintain the multimodal conversation ability, we also included a range of open-sourced uni/multimodal conversation datasets, denoted as $D_o$. We randomly shuffle and concatenate the single-turn samples from $D_p$ and $D_o$ to get a pseudo-multi-turn dataset $D_{pm}$. This dataset features multi-turn conversations but not necessarily preserving semantic coherence, simulating the scenarios in which the user may switch the topic within a conversation. To accommodate change of topic, we train the model to predict a \texttt{<switch>} token. We mix the collection of $D_o$, $D_p$, $D_{pm}$ together with $D_{tt}$ to serve as the final training dataset $D$.
For more details, please refer to \cite{huang2024dialoggen}.

\paragraph{Guarantee on Subject Consistency}
In multi-turn text-to-image, users may ask the AI system to edit a certain subject multiple times. Our goal is to ensure that the subjects generated across multiple conversational turns remain as consistent as possible. 
To achieve this, we add the following constraints in the `dialogue prompts' of the dialogue AI agent. For image generation that builds upon the images produced in previous turns, the transformed text prompts should satisfy the user's current demand while being altered as little as possible from the text prompts used for previous images.
Moreover, during the inference phase of a given conversation, we fix the random seed of the text-to-image model.  This approach significantly increases the subject consistency throughout the dialogue.

\subsection{System Efficiency Optimization}

\paragraph{Optimization in the Training Stage}

Due to the large number of model parameters in \name and the massive amount of image data required for training, we adopted ZeRO~\citep{rajbhandari2020zero}, flash-attention~\citep{dao2022flashattention}, multi-stream asynchronous execution, activation checkpointing, kernel fusion to enhance the training speed.

\paragraph{Optimization in the Inference Stage}

Deploying \name for the users is expensive, we adopt multiple engineering optimization strategies to improve the inference efficiency, including ONNX graph optimization, kernel optimization, operator fusion, precomputation, and GPU memory reuse.

\paragraph{Algorithmic Acceleration}
Recently, various methods have been proposed to reduce the inference steps of diffusion-based text-to-image models~\citep{luo2023latent, sauer2023adversarial, liu2023instaflow, xu2023ufogen, yin2023one}. We attempted to apply these methods to accelerate \name,
and the following problems arise:
\begin{enumerate}
    \item \textbf{Training Stability:} We observed adversarial training tends to collapse due to the unstable training scheme.
    \item \textbf{Adaptivity:} We found several methods results in models that cannot reuse the pre-trained plug-in modules or LoRAs.
    \item \textbf{Flexibility:} In our practice, the Latent Consistency Model is only suitable for low-step generation. Its performance deteriorates when the number of inference steps increases beyond a certain threshold. This limitation prevents us from flexibly adjusting the balance between generation performance and acceleration.
    \item \textbf{Training Cost:} Adversarial training introduces additional modules for training the discriminative model, which brings severe demand of extra GPU memory and training time. 
\end{enumerate}

Considering these problems, we choose Progressive Distillation~\citep{salimans2021progressive}. It enjoys stable training and allows us to smoothly trade-off between the acceleration ratio and the performance, offering us the cheapest and fastest way for model acceleration. To encourage the student model to accurately imitate the teacher model, we carefully tune the optimizer, classifier-free guidance, and regularization in the training process.

\section{Evaluation Protocol}
\label{sec:eval}

To holistically evaluate the generation ability of \name, we constructed a multi-dimensional evaluation protocol, which is composed of evaluation metrics, evaluation dataset construction, evaluation execution, and evaluation protocol evolution.

\subsection{Evaluation Metrics}
\paragraph{Evaluation Dimensions}
When determining the evaluation dimensions, we referenced existing literature and additionally invited professional designers and general users to participate in interviews to ensure that the evaluation metrics have both professionalism and practicality. Specifically, when evaluating the capabilities of our text-to-image models, we adopted the following four dimensions: text-image consistency, AI artifacts, subject clarity, and overall aesthetics. For results that raise safety concerns (such as involving pornography, politics, violence, or bloodshed), we directly mark them as \texttt{unacceptable}.

\paragraph{Multi-Turn Interaction Evaluation}
When evaluating the capabilities of the multi-turn dialogue interaction, we also assessed extra dimensions such as instruction compliance, subject consistency, and the performance of multi-turn prompt enhancement for image generation.

\subsection{Evaluation Dataset Construction}

\paragraph{Dataset Construction} 
We combine AI-generated and human-created test prompts to construct a hierarchical evaluation dataset with various difficulty levels. 
Specifically, we categorize the evaluation dataset into three difficulty levels - easy, medium, and hard - based on factors such as the richness of the text prompt content, the number of descriptive elements (main subject, subject modifiers, background descriptions, styles, etc.), whether the elements are common, and whether they contain abstract semantics (e.g. poems, idioms, proverbs).

Furthermore, due to the issues of homogeneity and long production cycles when creating test prompts with humans, we rely on LLMs to enhance the diversity and difficulty of the test prompts, rapidly iterate on prompt generation, and reduce manual labor.

\paragraph{Evaluation Dataset Categories and Distribution}
In the process of constructing hierarchical evaluation dataset, we analyzed the text prompts used by users when using the text-to-image generative models, and combined user interviews and expert designer opinions to cover functional applications, character roles, Chinese elements, multi-turn text-to-image generation, artistic styles, subject details, and other major categories in the evaluation dataset.

The different categories are further divided into multiple hierarchical levels. For example, the `subject details' category is further divided into subcategories like animals, plants, vehicles, and landmarks. For each subcategory, we maintain a prompt count of more than 30.

\begin{figure}
    \centering
    \includegraphics[width=0.85\textwidth]{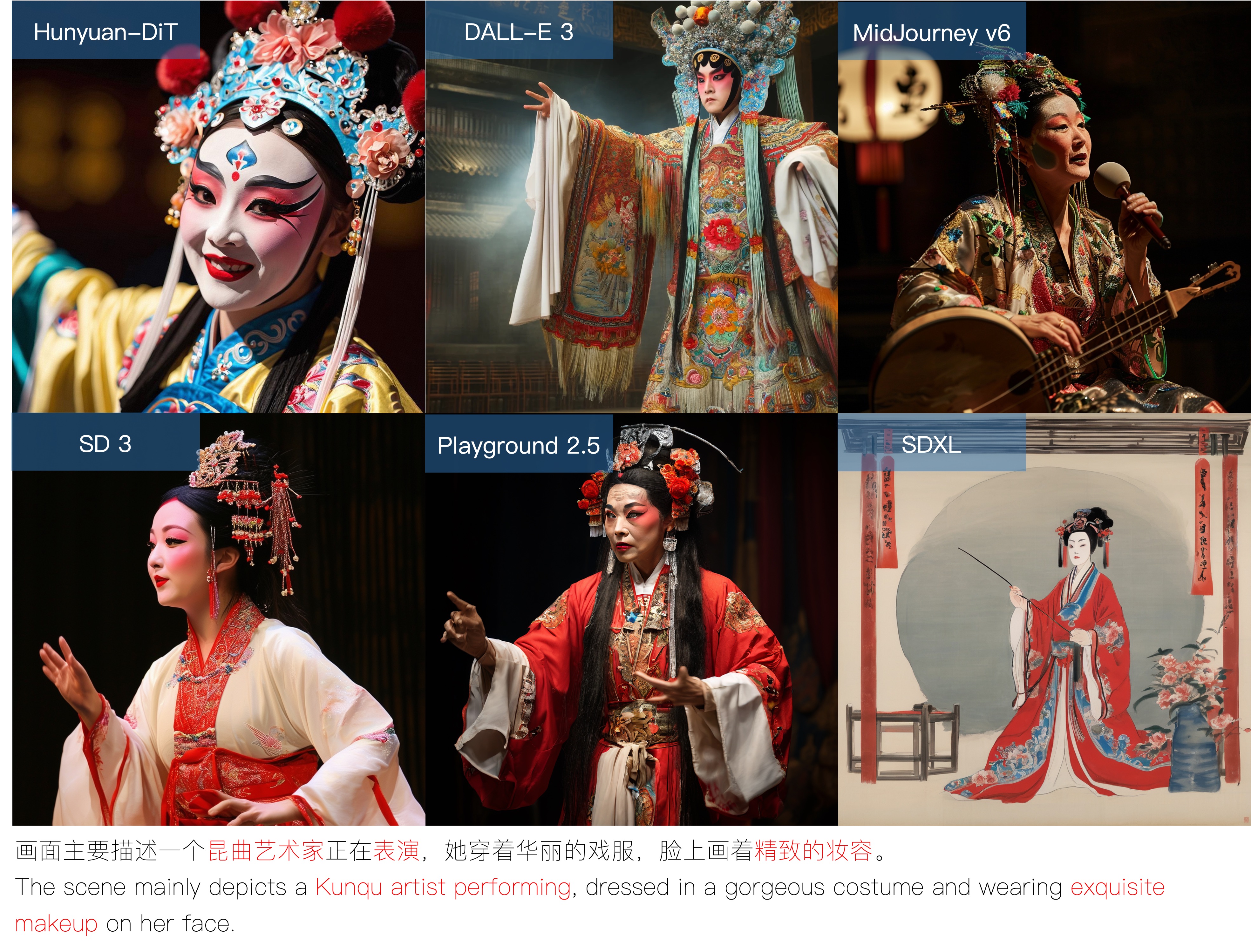}
    \caption{Qualitative comparison between \name and other SOTA models.}
    \label{fig:four_images}
\end{figure}

\begin{figure}
    \centering
    \includegraphics[width=0.85\textwidth]{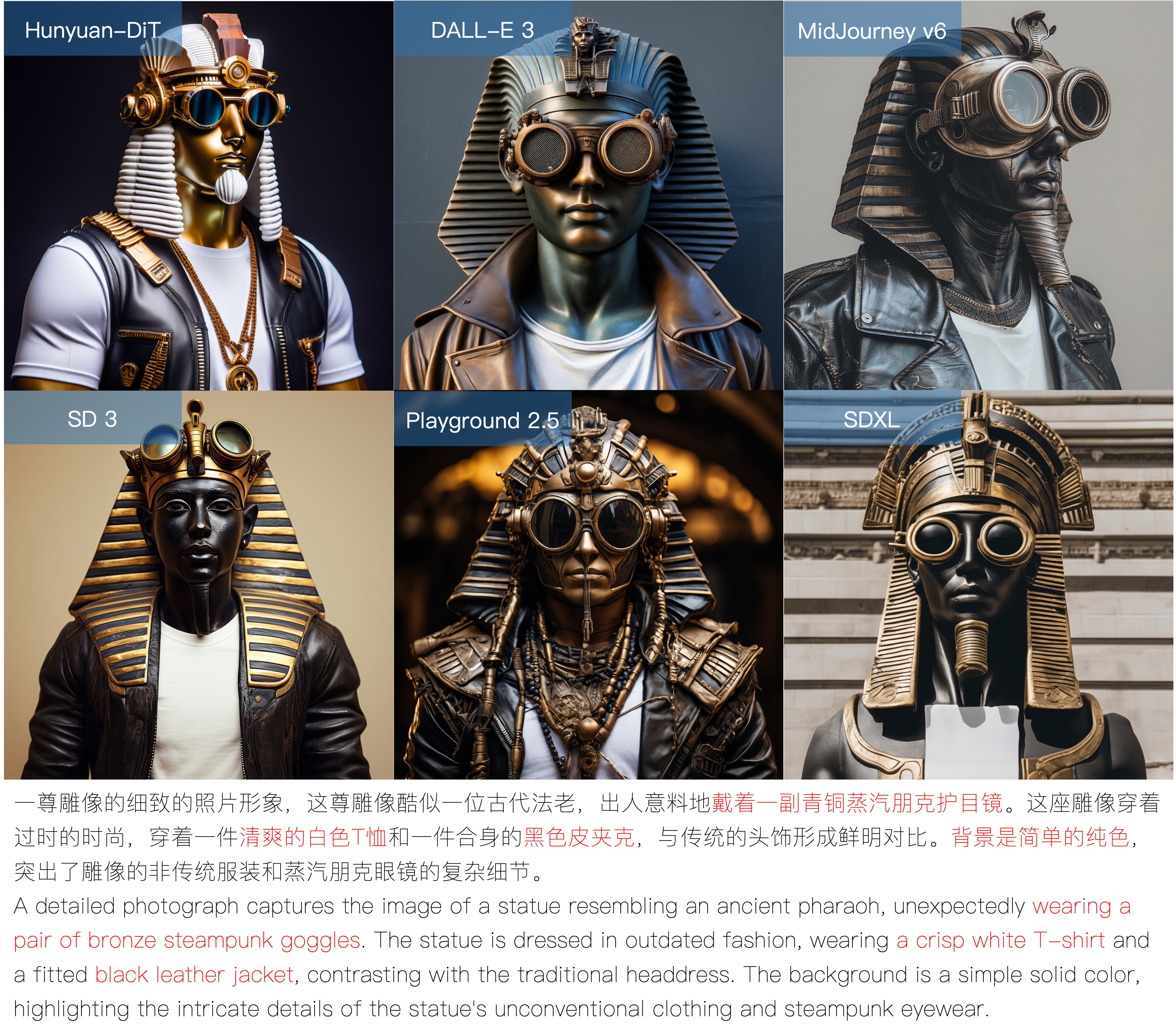}
    \caption{Qualitative comparison between \name and other SOTA models.}
    \label{fig:four_images}
\end{figure}

\begin{figure}
    \centering
    \includegraphics[width=0.85\textwidth]{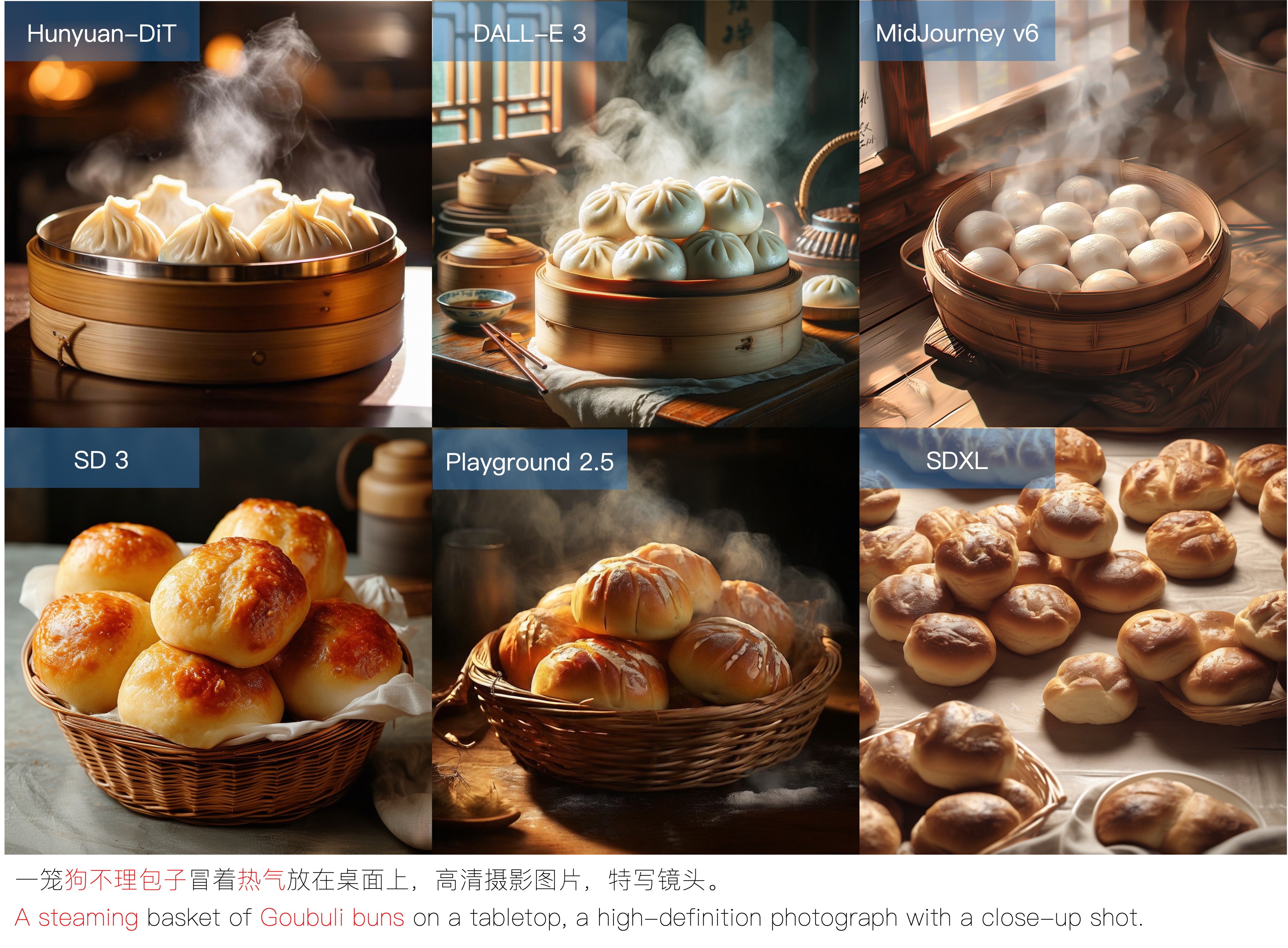}
    \caption{Qualitative comparison between \name and other SOTA models.}
    \label{fig:four_images}
\end{figure}

\subsection{Evaluation Execution}
\label{sec:eval_exec}
\paragraph{Evaluation Team}
The evaluation team consists of professional evaluators. They have rich professional knowledge and evaluation experience, allowing them to accurately execute the evaluation tasks and provide in-depth analysis. The evaluation team has more than 50 members.

\paragraph{Evaluation Process}
The evaluation process includes two stages: evaluation standard training and multi-person correction. 
In the evaluation standard training stage, we provide detailed training to the evaluators to ensure they have a clear understanding of the evaluation metrics and the tools.
In the multi-person correction stage, we have multiple evaluators independently evaluate the same set of images, then summarize and analyze the evaluation results to mitigate subjective biases among the evaluators.

Particularly, the evaluation dataset was structured in a 3-level hierarchical manner, with 8 level-1 categories and more than 70 level-2 categories. For each level-2 category, we have 30 - 50 prompts in the evaluation set. The evaluation set has more than 3,000 prompts in total. Specifically, our evaluation score is computed with the following steps:

\begin{enumerate}
    \item \textbf{Calculating Results for Individual Prompts:} For each prompt, we invite multiple evaluators to independently assess the images generated by the model. We then aggregate the evaluators' assessments and calculate the percentage of evaluators who consider the image to be acceptable. For example, if 10 evaluators are involved and 7 of them consider the image acceptable, the pass rate for that prompt is 70\%.
    \item \textbf{Calculating Level-2 Category Scores:} We classify the prompts into level-2 categories according to their contents. Each prompt within the same level-2 category has equal weight. For all the prompts under the same level-2 category, we calculate the average of their pass rates to obtain the score for that level-2 category. For example, if a level-2 category has 5 prompts with pass rates of 60\%, 70\%, 80\%, 90\%, and 100\%, the score for that level-2 category is (60\% + 70\% + 80\% + 90\% + 100\%) / 5 = 80\%.
    \item \textbf{Calculating Level-1 Category Scores:} Based on the level-2 category scores, we calculate the scores for the level-1 categories. For each level-1 category, we take the average of the scores of its subordinate level-2 categories to obtain the level-1 category score. For example, if a level-1 category has 3 level-2 categories with scores of 70\%, 80\%, and 90\%, the level-1 category score is (70\% + 80\% + 90\%) / 3 = 80\%.
    \item \textbf{Calculating the Overall Pass Rate:} Finally, we calculate the overall pass rate based on the weights of each level-1 category. Suppose there are 3 level-1 categories with scores of 70\%, 80\%, and 90\%, and weights of 0.3, 0.5, and 0.2 respectively, the overall pass rate would be 0.3 $\times$ 70\% + 0.5 $\times$ 80\% + 0.2 $\times$ 90\% = 79\%. The weights of the level-1 categories are determined by careful discussion with users, designers and experts, as shown in Table~\ref{tab:eval_weights}. 
\end{enumerate}

Through the above process, we can obtain the pass rates of the model at different category levels, as well as the overall pass rate, to comprehensively evaluate the model's performance.

\paragraph{Evaluation Result Analysis}
After evaluation, we conduct in-depth analysis of the results, including:
\begin{enumerate}
    \item Comprehensive analysis of the results for different evaluation metrics (text-image consistency, AI artifacts, subject clarity, and overall aesthetics) to understand the model's performance in various aspects.
    \item Comparative analysis of the model's performance on tasks of different difficulty levels to understand the model's capabilities in handling complex scenarios and abstract semantics.
    \item Identifying the model's strengths and weaknesses to provide directions for future optimization.
    \item Comparison with other state-of-the-art models.
\end{enumerate}

\subsection{Evaluation Protocol Evolution}
In the continuous optimization of the evaluation framework, we will consider the following aspects:
To improve our evaluation protocol to accommodate new challenges, we consider the following aspects: (1) introducing new evaluation dimensions;
(2) adding in-depth analysis in the evaluation feedback, such as the spots where the text-image inconsistency occurs, or precise markings of distortion locations; (3) dynamically adjusting the evaluation datasets; (4) improving evaluation efficiency by using machine evaluations.



\section{Results}

\subsection{Quantitative Evaluation}


\begin{figure}[!t]
\centering
\includegraphics[width=0.95\textwidth]{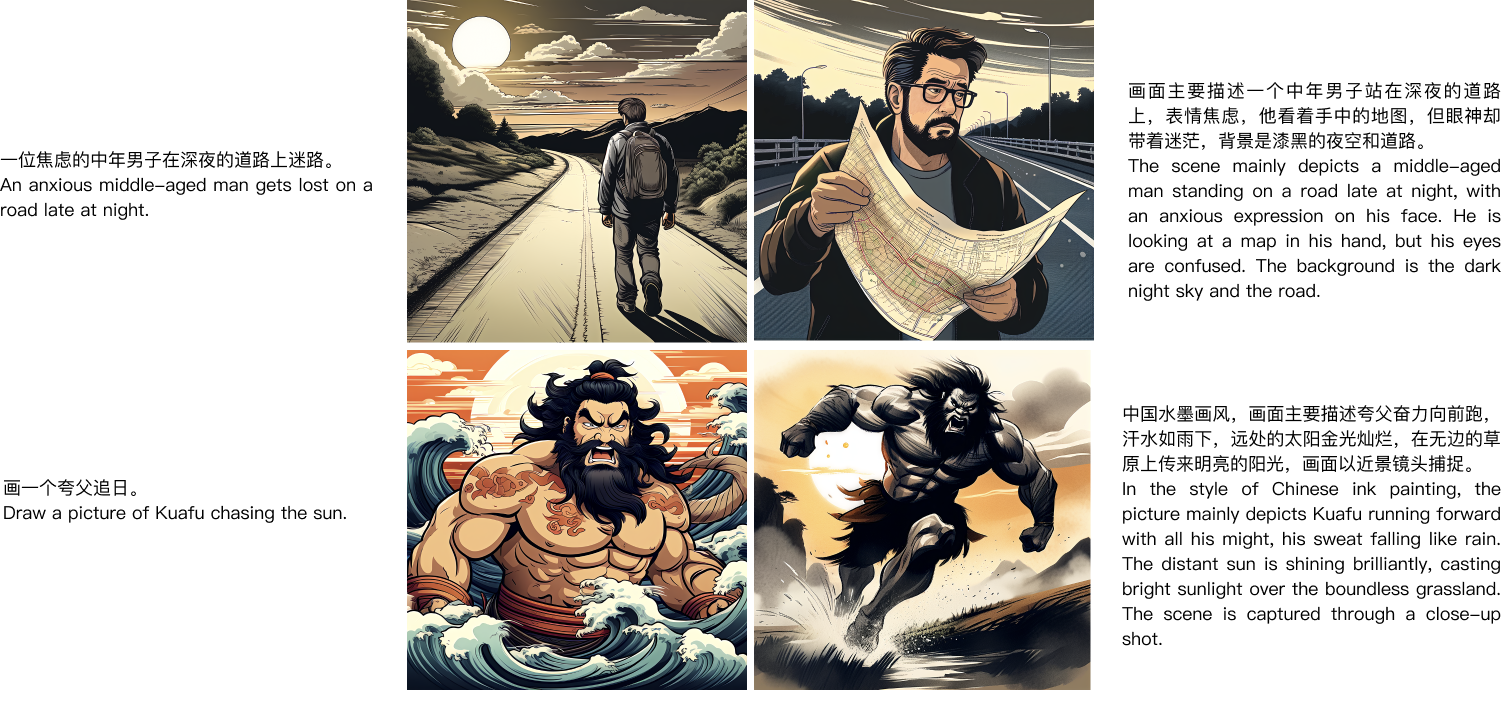}
\caption{
The effect of prompt enhancement. When it comes to simple abstract concept prompts, prompt enhancement with MLLM can effectively boost the consistency between generated images and their corresponding text descriptions.
}
\label{fig:prompt_enhancement}
\end{figure}

\paragraph{Comparison with State-of-the-Art}

We compared \name with state-of-the-art models, including both open-source models (Playground 2.5, PixArt-$\alpha$, SDXL) and closed-source models (DALL-E 3, SD 3, MidJourney v6). We follow the evaluation protocol in Section~\ref{sec:eval}. All the models are evaluated on four dimensions, including text-image consistency, the ability of excluding AI artifacts, subject clarity, and aesthetics. As depicted in Table~\ref{tab:sota}, \name achieves the best score on all the four dimensions compared with other open-source models. In comparison with closed-source models, \name can achieve similar performance to SOTA models such as MidJourney v6 and DALL-E 3 in terms of subject clarity and image aesthetics. In terms of the overall pass rate, \name ranks third among all models, better than existing open-source alternatives. \name has 1.5B parameters in total.


\subsection{Ablation Study}

\begin{table}[!t]
    \centering
    \resizebox{\linewidth}{!}{%
    \begin{tabular}{c|c|ccccc}
    \toprule
    \multirow{2}{*}{Type} & \multirow{2}{*}{Model}   & Text-Image   & Excluding   & Subject  & \multirow{2}{*}{Aesthetics (\%)} & \multirow{2}{*}{Overall (\%)} \\
    & & Consistency (\%) & AI Artifacts (\%) & Clarity (\%)
    \\
    \midrule
    \multirow{4}{*}{Open} & \textbf{Hunyuan-DiT} & \textbf{74.2}  & \textbf{74.3} & \textbf{95.4}  & \textbf{86.6}  & \textbf{59.0} \\
    
    & Playground 2.5~\citep{li2024playground} & 71.9 & 70.8  & 94.9 & 83.3 & 54.3 \\
    & PixArt-$\alpha$~\citep{chen2023pixart} & 68.3 & 60.9 & 93.2 & 77.5 & 45.5 \\
    & SDXL~\citep{podell2023sdxl} & 64.3 & 60.6 & 91.1 & 76.3 & 42.7 \\
    \midrule
    \multirow{3}{*}{Closed}& DALL-E 3~\citep{betker2023improving} & 83.9  & 80.3 & 96.5  & 89.4  & 71.0  \\
    & SD 3~\citep{esser2024scaling} & 77.1  & 69.3  & 94.6 & 82.5  & 56.7 \\
    & MidJourney v6~\citep{midjourney} & 73.5  & 80.2  & 93.5  & 87.2 & 63.3 \\ 
    \bottomrule
    \end{tabular}
    }
    \vspace{3mm}
    \caption{Comparison with other state-of-the-art models. \textbf{Bold} refers to the highest score in open-source models.}
    \label{tab:sota}
\end{table}

\paragraph{Experiment Setting} Following the setting in prior research~\cite{saharia2022photorealistic, balaji2022ediff}, 
we evaluate different variants of the models using the zero-shot Frechet Inception Distance (FID)~\citep{heusel2017gans} on MS COCO~\citep{lin2014microsoft} $256 \times 256$ validation dataset by generating 30,000 images from the prompts in the validation set. We also report the average CLIP~\citep{radford2021learning} score of these generated images to examine the correspondence between text prompts and images. These ablation studies are conducted on a smaller 0.7B diffusion transformer.

\begin{figure}[t]  
    \centering  
    \begin{subfigure}[b]{0.33\textwidth}  
        \includegraphics[width=\textwidth]{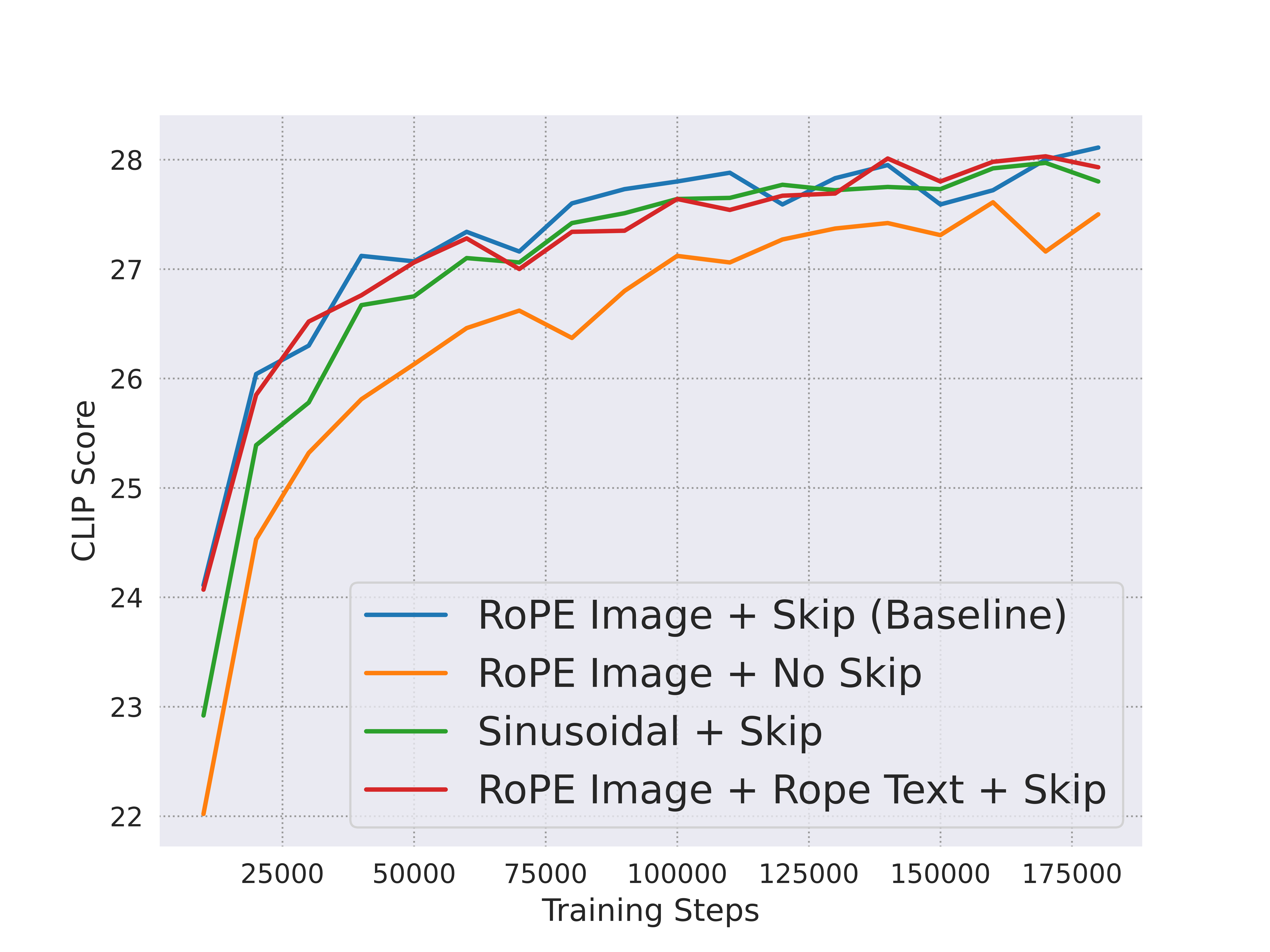} 
        \caption{CLIP Score}  
        \label{fig:abs_structure_clipscore}  
    \end{subfigure}  
    \begin{subfigure}[b]{0.33\textwidth}  
        \includegraphics[width=\textwidth]{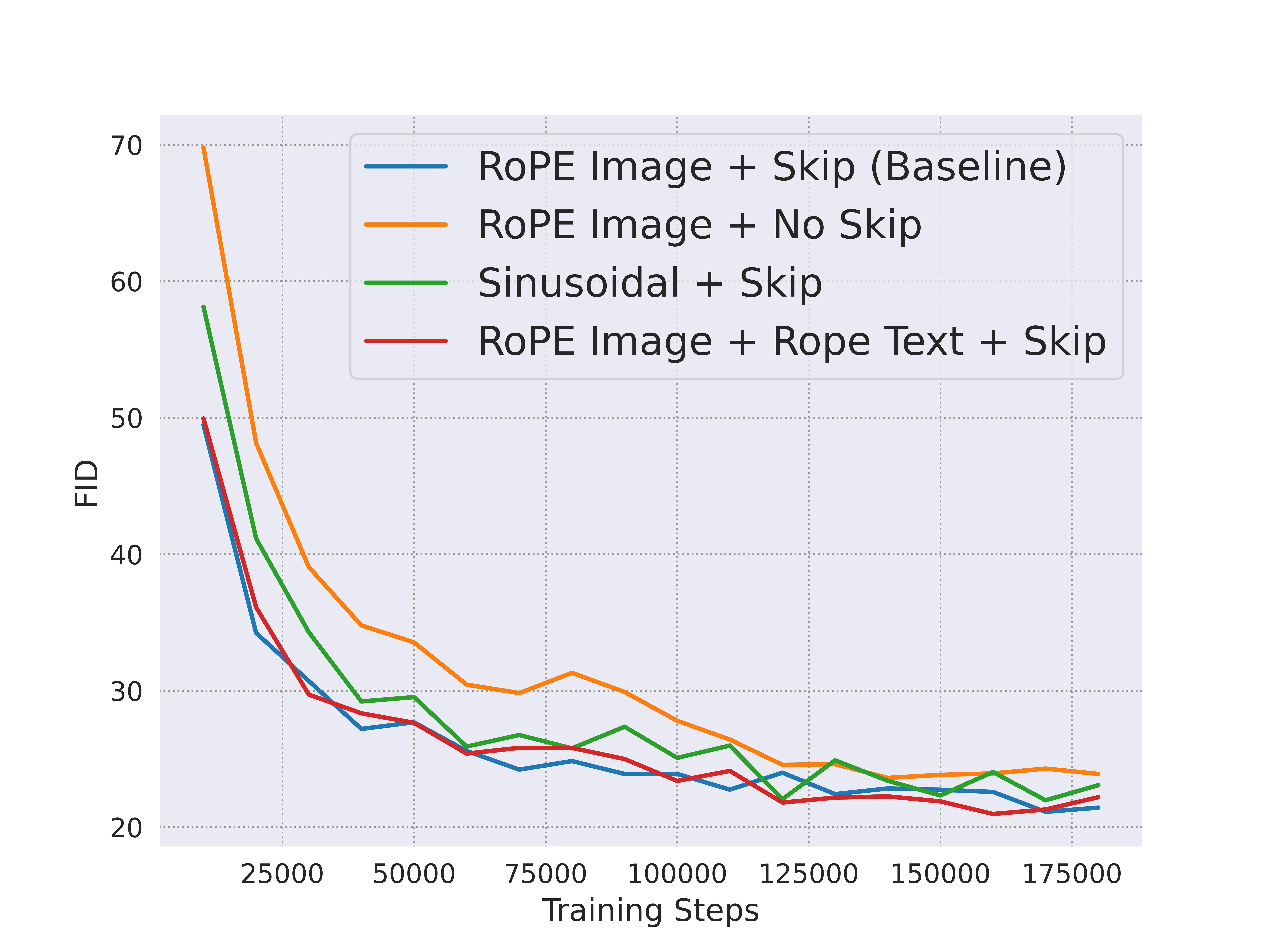} 
        \caption{FID}  
        \label{fig:abs_structure_fid}  
    \end{subfigure}  
    \caption{Ablation study on position encoding and model structure.}  
    \label{fig:abs_structure}  
\end{figure}

\paragraph{Effect of the Skip Module} Long skip connections are utilized to achieve feature fusion between symmetrically positioned encoding and decoding layers in U-Nets. We use Skip Modules in \name to mimic this design. As depicted in Figure.~\ref{fig:abs_structure},  we observed that removing long skip connection increases FID and decreases the CLIP score. 




\paragraph{Rotary Position Encoding (RoPE)} We compare sinusoidal position encoding~(the original position encoding in DiT~\citep{peebles2023scalable}) with RoPE~\citep{su2024roformer}. The results are shown in Figure.~\ref{fig:abs_structure} as well. We found RoPE position encoding outperformed the sinusoidal position encoding in most time of the training stage. Especially, we found RoPE accelerates the convergence of the model. We hypothesize that this is due to RoPE's ability to encapsulate both absolute and relative positional information.

We also evaluated the inclusion of one-dimensional RoPE position encoding in the text features, as shown in the Figure.~\ref{fig:abs_structure}. We found that adding RoPE position encoding to the text embeddings did not yield significant gains.

\begin{figure}[t]  
    \centering  
    \begin{subfigure}[b]{0.33\textwidth}  
        \includegraphics[width=\textwidth]{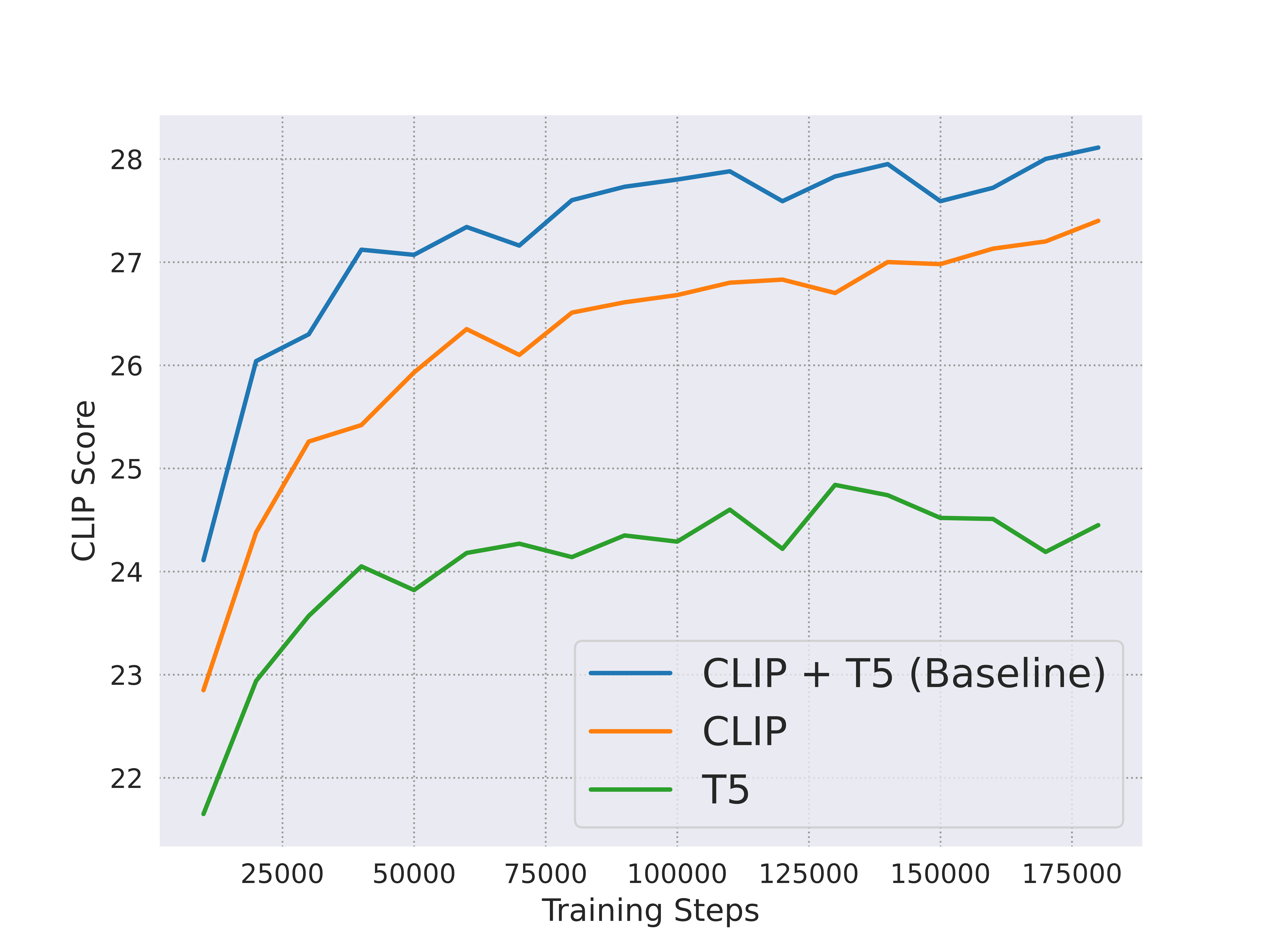} 
        \caption{CLIP Score}  
        \label{fig:abs_t5_clipsocre}  
    \end{subfigure}  
    \begin{subfigure}[b]{0.33\textwidth}  
        \includegraphics[width=\textwidth]{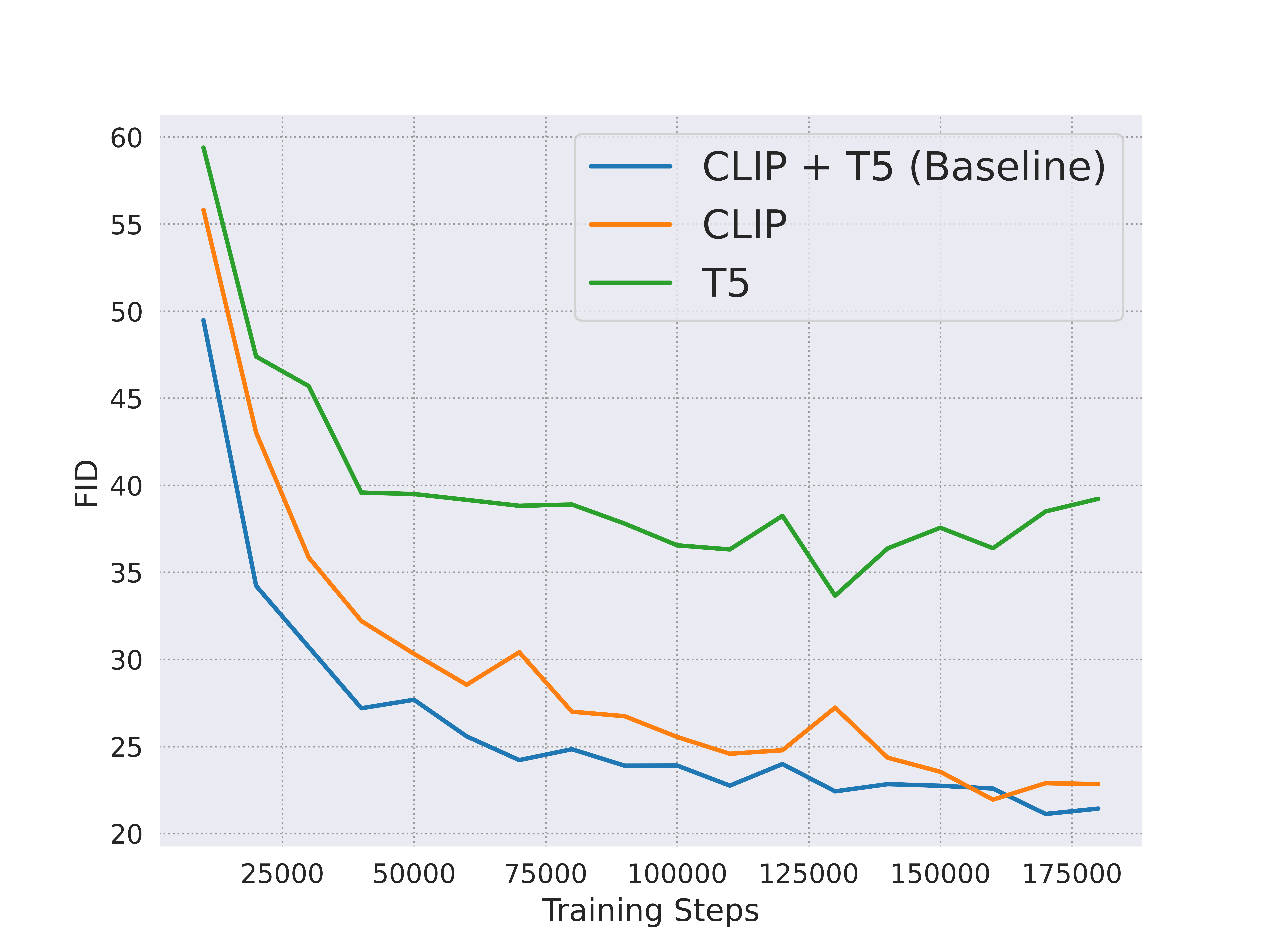} 
        \caption{FID}  
        \label{fig:abs_t5_fid}  
    \end{subfigure}  
    \caption{Ablation study on different schemes of text encoding.}  
    \label{fig:abs_t5}  
\end{figure}  

\begin{figure}[!h]  
    \centering  
    \begin{subfigure}[b]{0.33\textwidth}  
        \includegraphics[width=\textwidth]{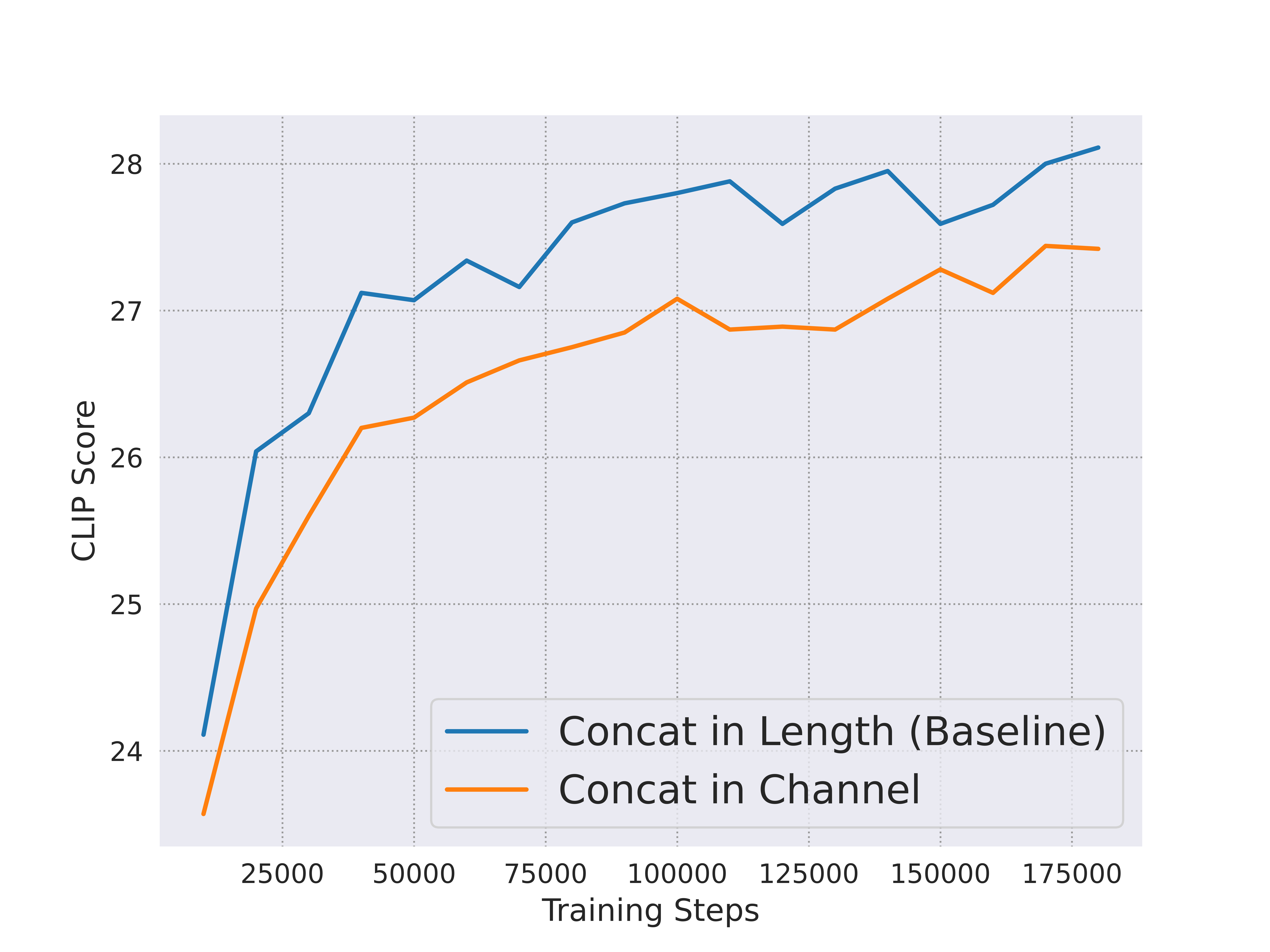} 
        \caption{CLIP Score}  
        \label{fig:abs_merge_text_encoder_clipscore}  
    \end{subfigure}  
    \begin{subfigure}[b]{0.33\textwidth}  
        \includegraphics[width=\textwidth]{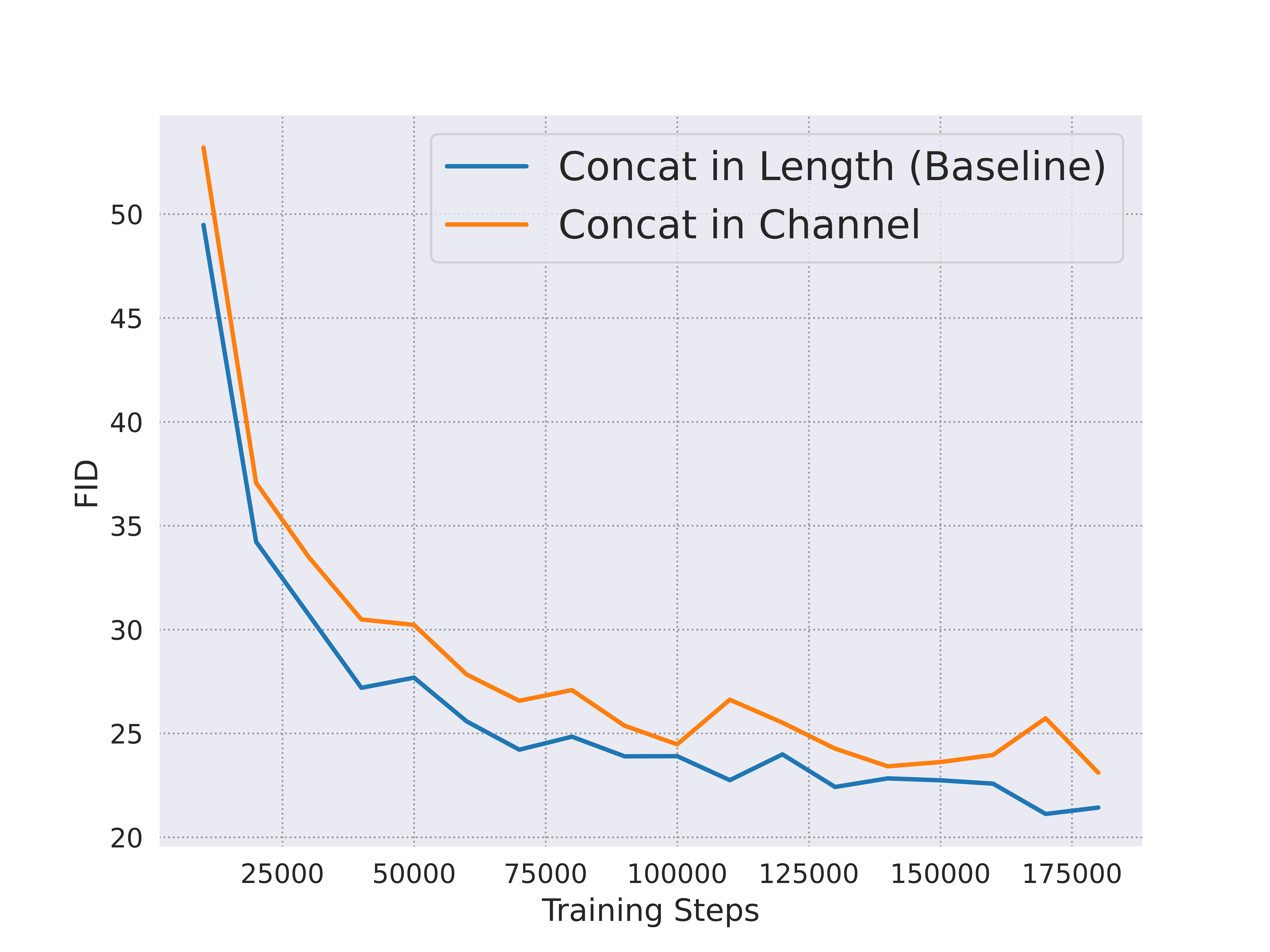} 
        \caption{FID}  
        \label{fig:merge_text_encoder_fid}  
    \end{subfigure}  
    \caption{Ablation study on concatenating features of the text encoders.}  
    \label{fig:abs_merge_text_encoder}  
\end{figure}  

\paragraph{Text Encoder} We evaluated three schemes for text encoding: (1) using our own bilingual (Chinese-English) CLIP alone, (2) using multilingual T5 alone, and (3) using both bilingual CLIP and multilingual T5. 
In Figure.~\ref{fig:abs_t5}, using CLIP encoder alone outperforms using multilingual T5 encoder alone. Moreover, combining the bilingual CLIP encoder with multilingual T5 encoder leverages both the efficient semantic capture ability of CLIP and the fine-grained semantic understanding advantage of T5, leading to a significantly enhanced FID and CLIP score.


We also explored two ways of concatenating features from CLIP and T5 in Figure.~\ref{fig:abs_merge_text_encoder}: merging along the channel dimension and merging along the length dimension. We found that concatenating the features of text encoders along the text length dimension yields superior performance. Our hypothesis is that, by concatenating along the text length dimension, the model can fully leverage the Transformer’s global attention mechanism to focus on each text slot. This facilitates a better understanding and integration of semantic information from different dimensions provided by T5 and CLIP.

\section{Conclusions}
In this report, we introduced the entire pipeline of building \namens, which is a text-to-image model with the ability to understand both English and Chinese. Our report elucidates the model design, data processing and the evaluation protocol of \namens. Combining these efforts from different aspects. \name reaches the top performance in Chinese-to-image generation among open-source models.
We hope \name can be a useful recipe for the community to train better text-to-image models. 
\clearpage
\bibliography{reference}
\bibliographystyle{plain}

\newpage
\appendix
\section{Additional Materials}
\begin{figure}[htbp]
    \centering
    \includegraphics[width=0.75\textwidth]{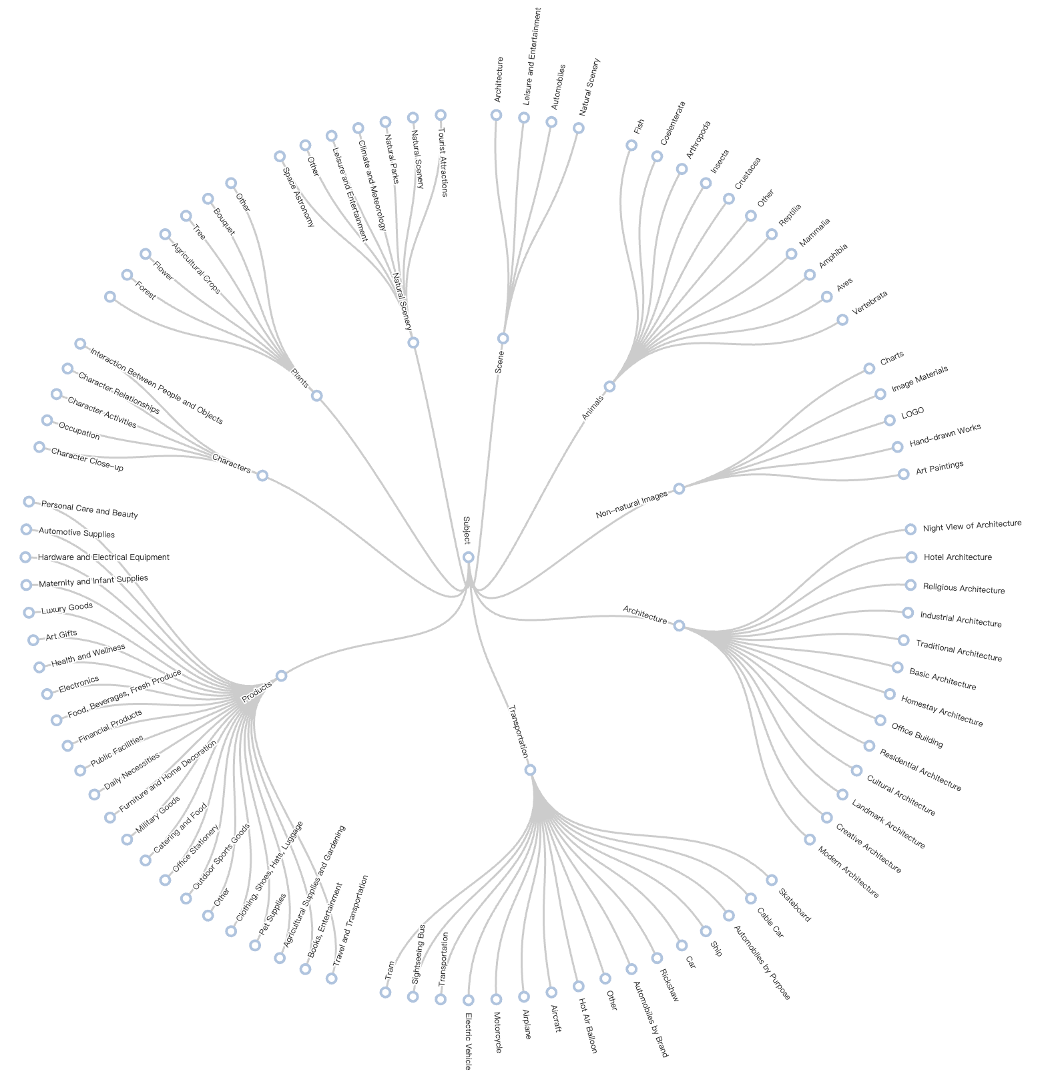}
    \caption{The hierarchy of subjects in our training data.}
    \label{fig:data_subject}
\end{figure}

\begin{figure}[htbp]
    \centering
    \includegraphics[width=0.75\textwidth]{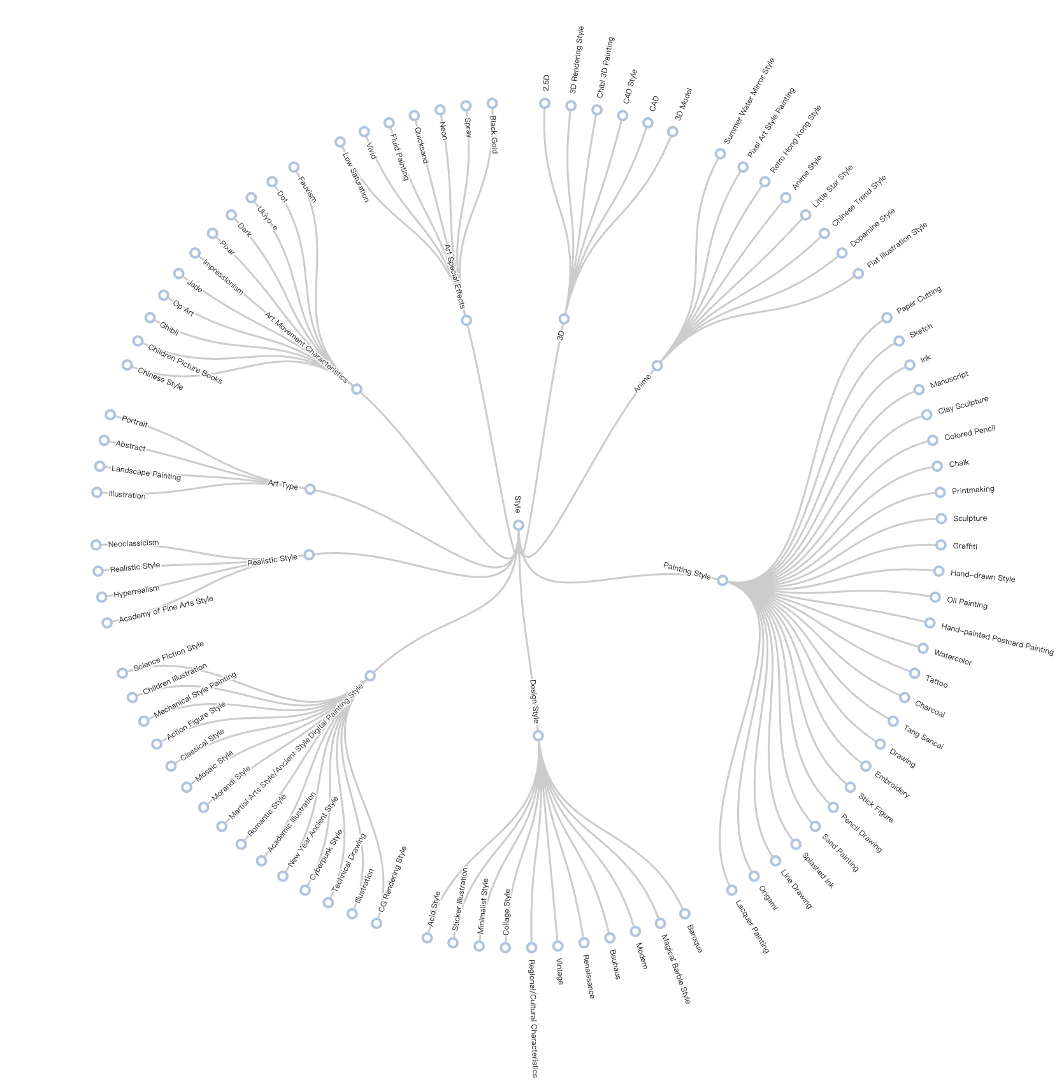}
    \caption{The hierarchy of styles in our training data.}
    \label{fig:data_style}
\end{figure}

\begin{figure}[htbp]
    \centering
    \includegraphics[width=0.8\textwidth]{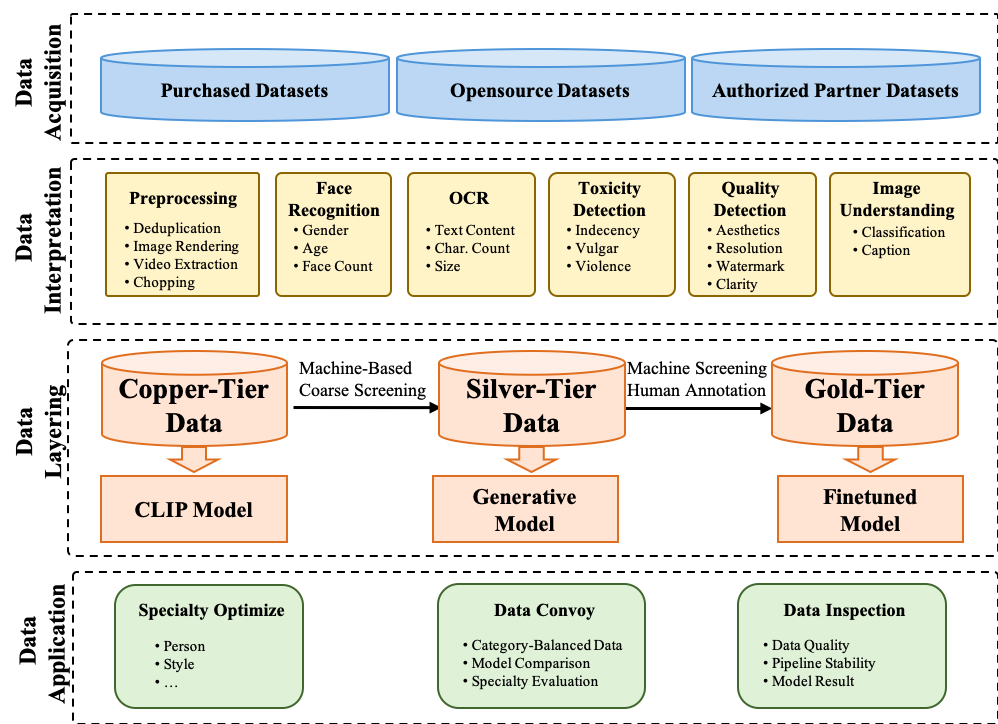}
    \caption{Illustration of our whole data pipeline.}
    \label{fig:data_pipeline}
\end{figure}

\begin{figure}[htbp]
    \centering
    \includegraphics[width=0.85\textwidth]{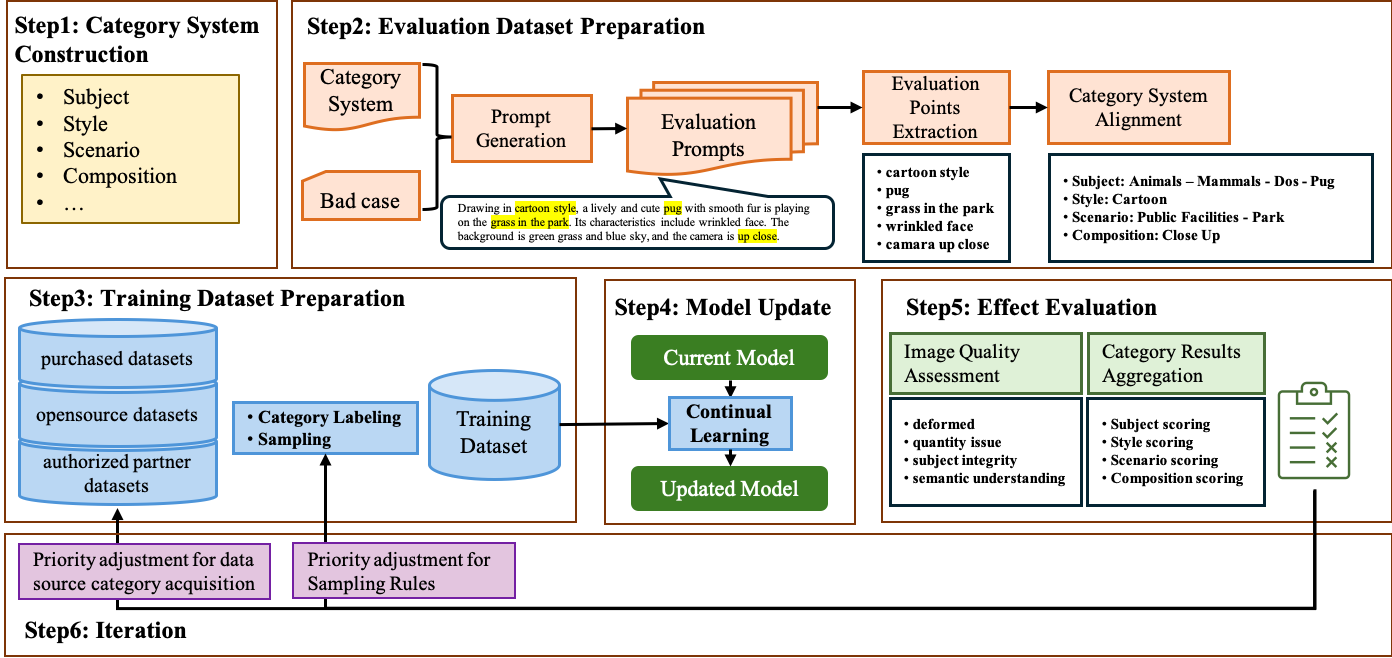}
    \caption{Illustration of our `data convoy' mechanism. }
    \label{fig:data_convoy}
\end{figure}

\clearpage
\begin{table}[htbp]
    \centering
    \begin{tabular}{lc}
    \toprule
    Categories & Weights \\ 
    \midrule
    Functional Images & 5\% \\
    Human Characters & 20\% \\
    Iconic Imagery & 5\% \\
    Chinese Elements & 10 \% \\ 
    Artistic Styles & 20 \%  \\
    Spatial Composition & 10 \% \\ 
    Subject and Details & 20 \%   \\
    \bottomrule
    \end{tabular}
    \caption{Weights of different categories in our evaluation protocol. Note that the weights are summed to 90\% because we put 10\% for multi-turn text-to-image generation when evaluating our own model internally. For comparison with SOTA, we only consider the cateogires in the table.}
    \label{tab:eval_weights}
\end{table}




\end{document}